\documentclass[default]{sn-jnl}

\usepackage{times}
\usepackage{epsfig}
\usepackage{graphicx}
\usepackage{adjustbox}
\usepackage{amsmath}
\usepackage{amssymb}
\usepackage{mathtools}
\usepackage[normalem]{ulem}

\usepackage{caption}
\usepackage{subcaption}
\usepackage{array, multirow}
\usepackage{booktabs}

\newcommand{\eg}{e.g., }
\newcommand{\ie}{i.e., }
\newcommand{\etal}{et al. }
\newcommand{\noi}{\noindent }

\DeclarePairedDelimiter{\nround}\lfloor\rceil

\DeclareMathOperator*{\argmax}{arg\,max}
\DeclareMathOperator*{\argmin}{arg\,min}
\DeclareMathOperator{\spn}{span}

\DeclareMathOperator{\sgn}{sgn}

\DeclareMathOperator*{\median}{Median}

\newcommand{\Dim}{\text{Dim}}

\renewcommand{\vec}[1]{\boldsymbol{#1}}
\newcommand{\matr}[1]{\boldsymbol{#1}}
\newcommand{\coord}[1]{\boldsymbol{#1}}
\newcommand{\tensor}[1]{\boldsymbol{#1}}
\newcommand{\Reals}{\mathbb{R}}
\newcommand{\Naturals}{\mathbb{N}}

\newcommand{\disparity}{\coord{d}}
\newcommand{\pixel}{\coord{p}}

\newcommand{\dotprod}[2]{\left\langle#1,#2\right\rangle}
\newcommand{\costMeasure}{\rho}
\newcommand{\patchingOp}[2][ ]{P_{\tensor{#2}}^{#1}}

\newcommand{\expec}[2]{\mathrm{E}_{#1}\left[ #2 \right]}

\newcommand{\bigO}[1]{\mathcal{O}(#1)}

\jyear{2023}

\begin{document}

\title[Generalized Closed-form Formulae for Feature-based Subpixel Alignment]{Generalized Closed-form Formulae for Feature-based Subpixel Alignment in Patch-based Matching}

\author*[1,2]{\fnm{Laurent Valentin} \sur{Jospin}}\email{laurent.jospin@epfl.ch}
\author[3]{\fnm{Hamid} \sur{Laga}}\email{H.Laga@murdoch.edu.au}
\author[4]{\fnm{Farid} \sur{Boussaid}}\email{farid.boussaid@uwa.edu.au}
\author[2]{\fnm{Mohammed} \sur{Bennamoun}}\email{mohammed.bennamoun@uwa.edu.au}

\affil[1]{\orgdiv{Faculté ENAC}, \orgname{École Polytechnique Fédérale de Lausanne}, \orgaddress{\street{Station 15}, \city{Lausanne}, \postcode{CH-1015}, \state{VD}, \country{Switzerland}}}

\affil[2]{\orgdiv{School of Computer Science and Software Engineering}, \orgname{University of Western Australia}, \orgaddress{\street{35 Stirling Highway}, \city{Perth}, \postcode{6009}, \state{WA}, \country{Australia}}}

\affil[3]{\orgdiv{School of Information Technology}, \orgname{Murdoch University}, \orgaddress{\street{90 South Street}, \city{Murdoch}, \postcode{6150}, \state{WA}, \country{Australia}}}

\affil[4]{\orgdiv{School of Electrical Engineering}, \orgname{University of Western Australia}, \orgaddress{\street{35 Stirling Highway}, \city{Perth}, \postcode{6009}, \state{WA}, \country{Australia}}}

\abstract{Patch-based matching is a technique meant to measure the disparity between pixels in a source and target image and is at the core of various methods in computer vision. When the subpixel disparity between the source and target images is required, the cost function or the target image has to be interpolated. While cost-based interpolation is easier to implement, multiple works have shown that image-based interpolation can increase the accuracy of the disparity estimate. In this paper we review closed-form formulae for subpixel disparity computation for one dimensional matching, e.g., rectified stereo matching, for the standard cost functions used in patch-based matching. We then propose new formulae to generalize to high-dimensional search spaces, which is necessary for unrectified stereo matching and optical flow. We also compare the image-based interpolation formulae with traditional cost-based formulae, and show that image-based interpolation brings a significant improvement over the cost-based interpolation methods for two dimensional search spaces, and small improvement in the case of one dimensional search spaces. The zero-mean normalized cross correlation cost function is found to be preferable for subpixel alignment. A new error model, based on very broad assumptions is outlined in the Supplementary Material to demonstrate why these image-based interpolation formulae outperform their cost-based counterparts and why the zero-mean normalized cross correlation function is preferable for subpixel alignement.}


\maketitle

\section{Introduction}
\label{sec:introduction}


The goal of image-based patch matching is to find, for a given patch in a source image $\tensor{I}_s$, the location of the same (or corresponding) patch in one or multiple target images $\tensor{I}_t$. The change in coordinates between the patch in $\tensor{I}_s$ and its corresponding patch in $\tensor{I}_t$ is called \emph{disparity} and is denoted by  $\disparity$. Estimating the disparity $\disparity$ is a fundamental problem and a building block to many  computer vision applications such as (multi-view) stereo vision~\cite{4270273, 4620119}, optical flow estimation \cite{10.1007/978-3-642-40602-7_37}, non-rigid registration \cite{sur:hal-02862808}, velocimetry \cite{Nobach2005}, photogrammetry and remote sensing \cite{rs12040696, debella2010sub}. Although it has been extensively investigated, using either traditional methods~\cite{4270273} or modern deep learning-based approaches~\cite{laga2020survey}, achieving fine subpixel-level accurate matching remains a challenging problem.

\begin{figure}[t]
    \centering
    
    \begin{subfigure}[t]{0.7\textwidth}
        \includegraphics[width=\textwidth]{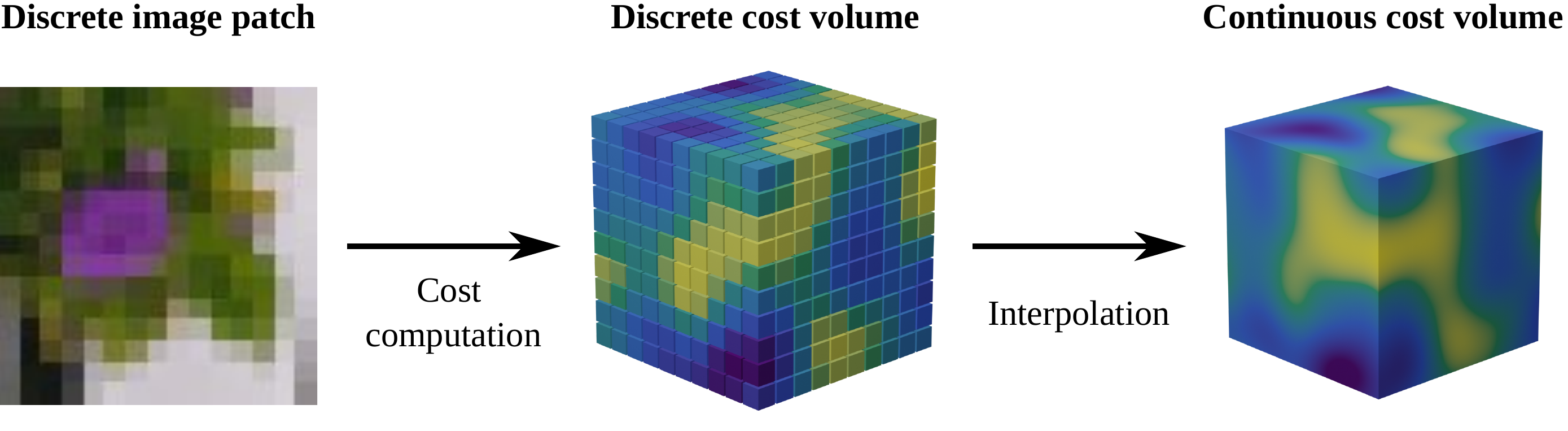}
        \caption{Cost interpolation}
        \label{fig:approaches:costInterp}
    \end{subfigure}
    
    \vspace{5pt}
    
    \begin{subfigure}[t]{0.7\textwidth}
        \includegraphics[width=\textwidth]{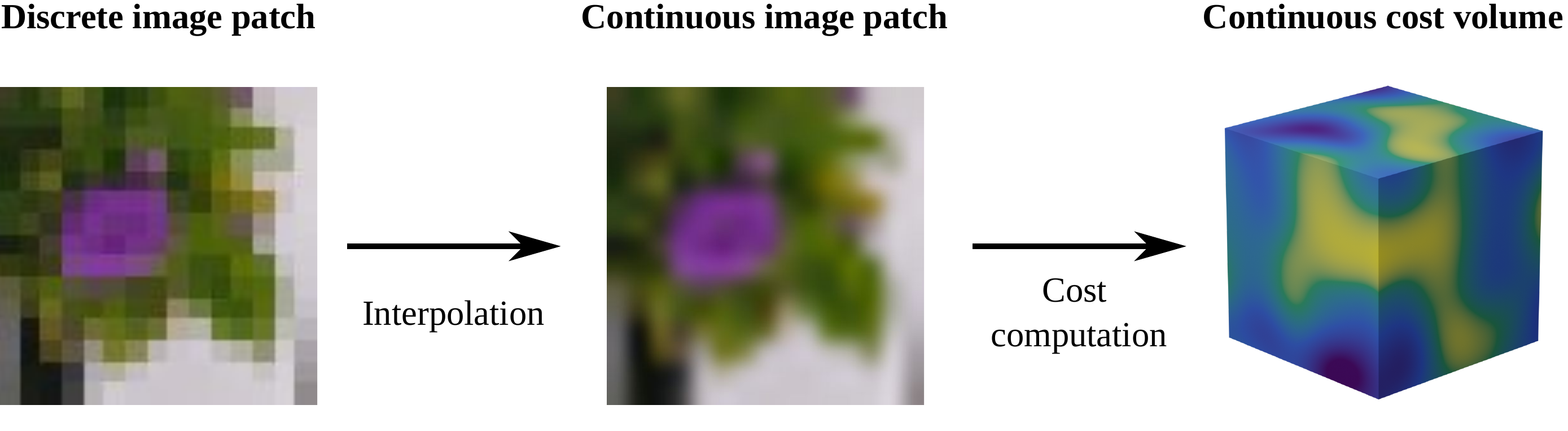}
        \caption{Image interpolation}
        \label{fig:approaches:imgInterp}
    \end{subfigure}
    
    \caption{In patch-based matching, two approaches can be used to achieve subpixel precision: computing the cost volume at discreet intervals and then interpolating (\subref{fig:approaches:costInterp}) or interpolating in the image space to directly compute a continuous cost volume (\subref{fig:approaches:imgInterp}). The former approach is generally preferred for its simplicity, even though the latter is more accurate and suffer from less biases.}
    \label{fig:interp_approaches}
\end{figure}

\begin{figure}[t]
    \centering
    
    \includegraphics[width=0.7\textwidth]{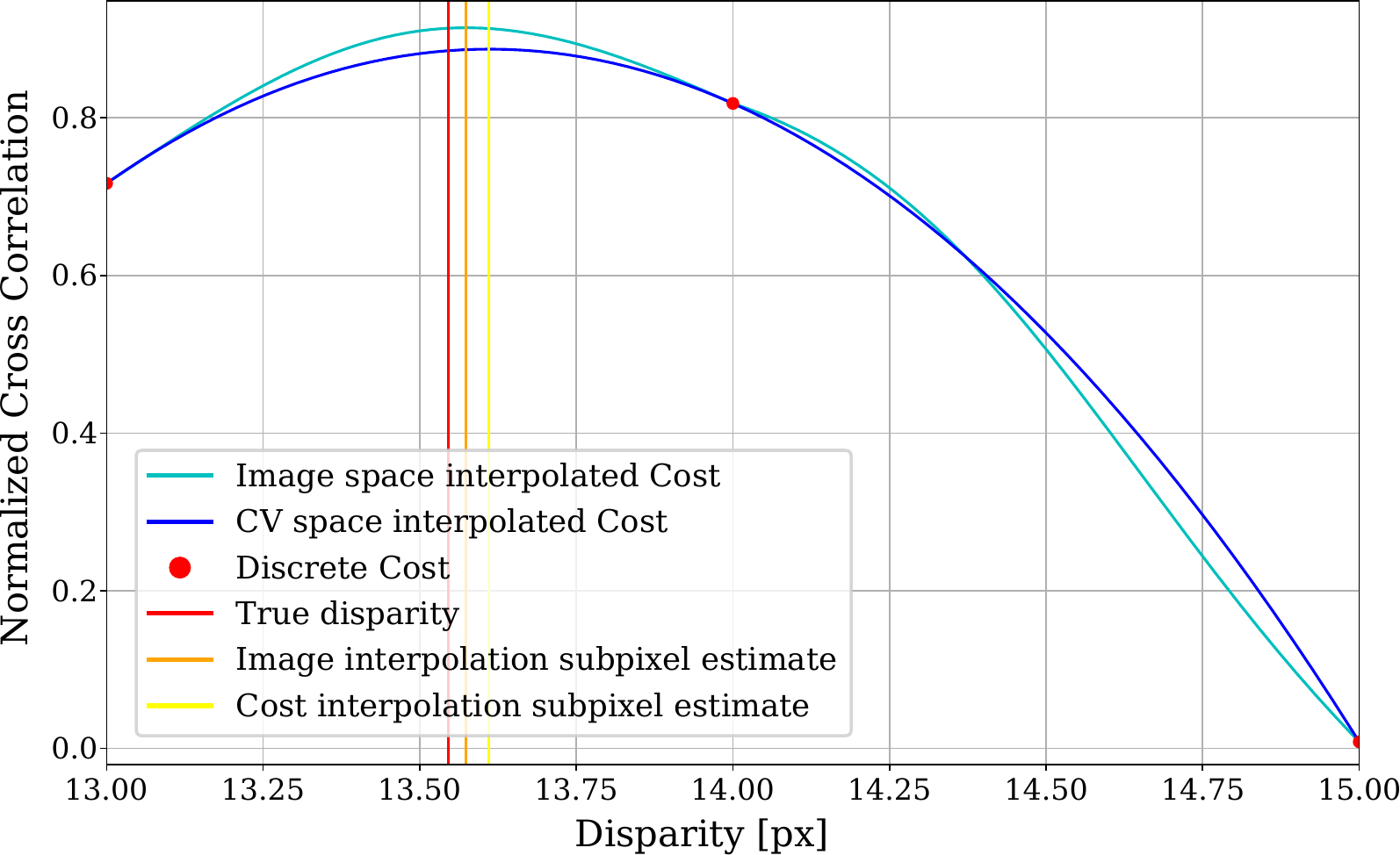}
    
    \caption{Different cost curves for different cost volume interpolation methods: image space vs cost volume space (example extracted from a real image).}
    \label{fig:cv_interpolation_methods}
\end{figure}

The most generic method to achieve subpixel accuracy is to construct a continuous cost volume $C(\pixel,\disparity)$ which measures, for every pixel $\pixel$ and disparity value $\disparity$,  how well a patch centered at  $\pixel$ on the source image  matches a patch located at $\pixel + \disparity$  on the target image. Two approaches can be used to build the continuous cost volume (see Fig.~\ref{fig:interp_approaches}). The continuous cost volume can be obtained by first evaluating $C$ at discrete intervals of $\pixel$ and $\disparity$, usually at one pixel increments, and then interpolating the resulting discrete cost volume, obtaining a continuous volume from which a finer, fractional estimate of the disparity $\hat{\disparity}$ can be estimated (Fig.~\ref{fig:approaches:costInterp}). Alternatively, the interpolation can  be performed in  the image space where a continuous version of the cost volume is computed from a continuous version of the source and target images (Fig.~\ref{fig:approaches:imgInterp}). In practice, the former, \ie cost volume interpolation, is often the preferred method since it results in simple and generic closed form expressions. However, the latter, \ie image interpolation, has been shown to result in more accurate disparity estimates than the cost volume interpolation-based methods \cite{4409212, 6460147, 1541350, Delon2007} (see Fig.~\ref{fig:cv_interpolation_methods}). This is because image signals are expected to have a lower bandwidth than cost curves, therefore, they vary more slowly and are better suited for interpolation. This is a major reason why image interpolation is more effective than cost interpolation. This is explained by the Bernstein inequality in signal processing; see section I in the Supplementary Material for more details. 

Some papers have already proposed closed form solutions for image based interpolation for some specific cost functions like the Sum of Square Differences (SSD) \cite{8960959, LMedSqFlow} or the Normalized Cross Correlation (NCC) \cite{1541350, Delon2007}. To the best of our knowledge, no work has proposed a formula for the Sum of Absolute Differences (SAD) cost function. Some of these formulae do generalize to n-dimensional patch based matching \cite{8960959, LMedSqFlow}, but only for the Sum of Square Differences (SSD) cost function. A generic closed form solution for image-based subpixel refinement applicable to all common cost functions and applicable to higher dimensions is still missing in the literature.

In this paper, we focus on image-based interpolation and show that closed-form formulae can be derived in a generic manner when using common dissimilarity measures such as NCC, SSD, and the SAD. Our formulation is also easy to generalize to cases where the disparity is two dimensional, like \eg for non-rectified stereo images and for optical flow estimation, and tri- or even higher-dimensional as required in applications such as medical imaging \cite{RUIZALZOLA2002143} or fluid dynamics~\cite{Discetti2012}, which are based on volumetric data~\cite{LU2019157}. The contributions of this paper are as follows;

\begin{itemize}
    \item We propose a novel closed-form formula for image-based interpolation for 1D sub-pixel disparity refinement when using the SAD as a dissimilarity function.
    
    \item We also demonstrate that although there is no closed-form formula for the SAD for two or more dimensions, a simplex-like iterative algorithm \cite{10.1145/990308.990310} is guaranteed to lead to an exact solution in a finite number of iterations.
    
    \item We propose novel closed-form formulae for image-based interpolation for 2D, 3D and higher dimensional sub-pixel disparity refinement when using the NCC or ZNCC dissimilarity functions.  
    
    \item Additionally, our formulation is generic as it can easily be adapted to new cost or score metrics. As it is formulated in tensor notation, it can also be integrated into pipelines based on tensor processing libraries (\eg deep learning).
    
    \item We compare the subpixel accuracy of the proposed and pre-existing formulae and show that the proposed formulae, especially the ones based on ZNCC, bring a significant improvement over the cost-based methods when the search space is two dimensional, and small improvement in the case of one dimensional search spaces. The ZNCC cost function is found to yield the best estimates of all considered cost functions. We support these results with a mathematical error model for subpixel disparity estimation, discussed in the Supplementary Material. This model is based on the assumption that images are bounded and band-limited signals, which allows for precise predictions and validations of the subpixel accuracy under controlled conditions. This framework is crucial for understanding the limitations and potential accuracy of our disparity estimation methods. It explains why these image-based interpolation formulae outperform their cost-based counterparts and why the ZNCC function is preferable for subpixel alignement.

\end{itemize}

\noi As the underlying problem is generic for any patch-based matching model, the proposed closed-form formulae can benefit a wide range of applications. For example, sparse patch-based matching is very important in robotics, for Simultaneous Localization an Mapping (SLAM) and related applications, where a good subpixel estimate of a reference patch position is required for a good estimate of a robot location \cite{drones3030069}. Patch-based matching is also used in dense stereo reconstruction where a good subpixel accuracy is required to avoid visual artifacts, \eg for augmented reality applications \cite{Orts-EscolanoSergio2016Holoportation}.
In addition to the traditional stereo matching methods, which are still widely adopted, especially for lightweight embedded systems~\cite{Keselman_2017_CVPR_Workshops} and remote sensing \cite{rs12040696}, the proposed formulae can benefit the modern deep learning models for stereo matching and optical flow estimation~\cite{laga2020survey} which use explicit feature matching modules such as cross correlation layers
\cite{semi-dense-stereo2019, Knobelreiter2021}. It can also benefit self-supervised depth estimation where (multiview) stereo images are used as supervisory signals~\cite{Zhang_2018_ECCV}. 


The remainder of the paper is organized as follows; Section~\ref{sec:related_work} reviews the related work. Section~\ref{sec:notations} sets the  notation used throughout the paper and provides the important definitions. Section~\ref{sec:interpFormulae} derives, for various cost functions, closed form formulae for unidimensional fractional disparity reconstruction based on interpolation in the image (feature) space  (Section~\ref{sec:imageInterpFormulae}), compared to the traditional cost based interpolation functions (Section~\ref{sec:costInterpFormulae}). Section~\ref{sec:formulae_higher_dims} discusses how the approaches presented in Section~\ref{sec:interpFormulae} can be generalized to higher dimensions. Section~\ref{sec:algo_complexity} discusses the algorithmic complexity of the different formulae and algorithms presented in the previous sections. Our experimental results, presented in Section~\ref{sec:experimental}, show that the proposed formulae bring an improvement in terms of accuracy over cost volume interpolation. The benefits are especially important in the case of optical flow estimation. Section~\ref{sec:conclusion} concludes the paper and discusses directions for future research. The Supplementary Material provides additional proofs and derivations of the formulae presented in this paper. It also provides additional discussions on the error model of the proposed formulae. We provide an open source library implementing the different formulae presented in this paper at \url{https://github.com/french-paragon/LibStevi}.


\section{Related work}

\begin{figure}[t]
    \centering
    \begin{subfigure}[t]{0.45\textwidth}
        \includegraphics[height=3.8cm]{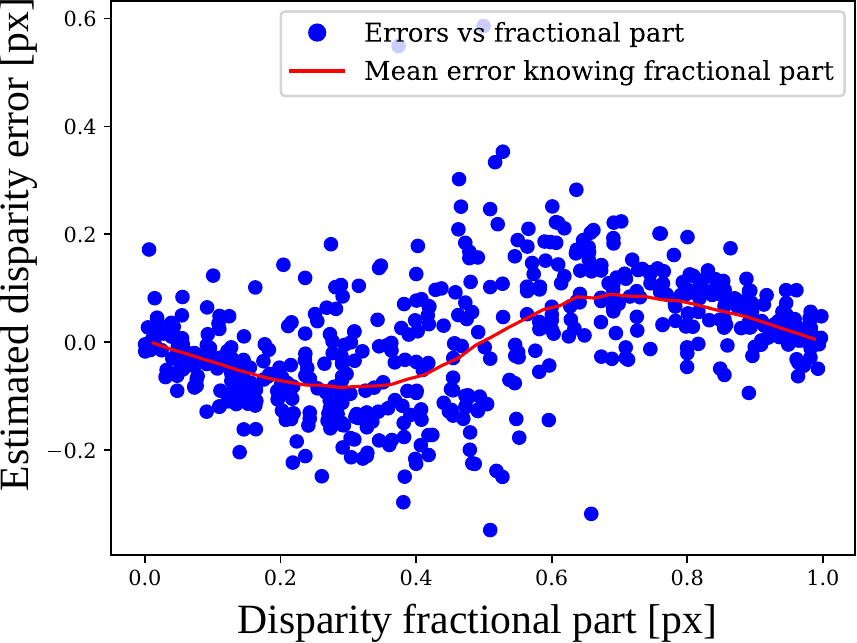}
        \caption{Cost based interpolation}
        \label{fig:fract_part_correllation:cost}
    \end{subfigure}
    \hfill
    \begin{subfigure}[t]{0.45\textwidth}
        \includegraphics[height=3.8cm]{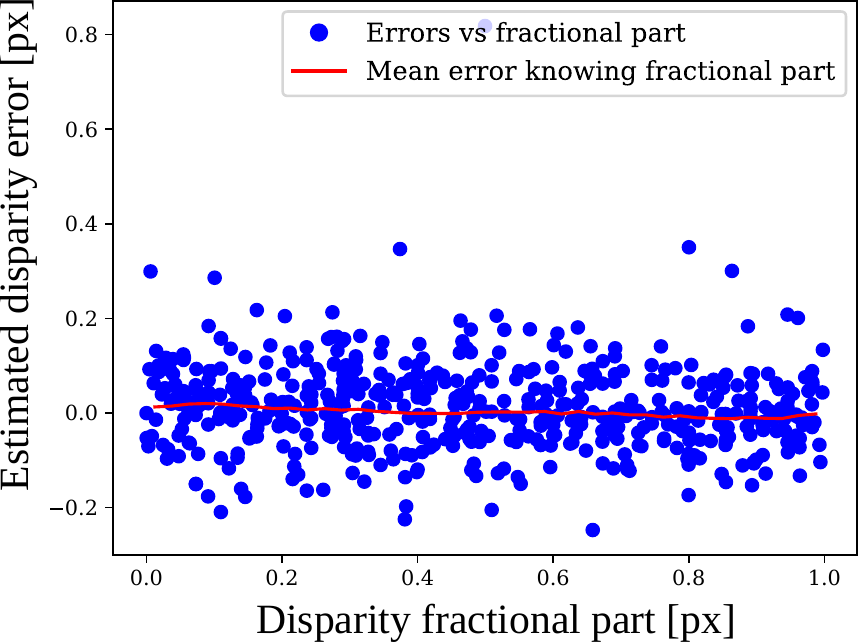}
        \caption{Image based interpolation}
        \label{fig:fract_part_correllation:image}
    \end{subfigure}
    
    \caption{The pixel-locking effect, observed in cost volume-based interpolation methods (\subref{fig:fract_part_correllation:cost}), occurs when the expected disparity error varies with the disparity fractional part. However, it does not occur with image-based interpolation (\subref{fig:fract_part_correllation:image}).}
    \label{fig:fract_part_correllation}
\end{figure}

\label{sec:related_work}
In this section, we review the different closed form solutions that have been used to achieve subpixel accuracy for disparity estimation. We classify existing closed form solutions based on whether the interpolation is done using the cost volume or the image space. We then discuss the methods for generalizing the closed form solutions to multidimensional disparity.

\vspace{6pt}
\noindent\textbf{Cost volume interpolation.} Cost volume interpolation is usually performed by fitting a parabola~\cite{Shimizu2005,Cheng_2013}, or in some cases a pair of equiangular lines forming a triangular function,  around each discrete disparity value. This approach is problematic since it leads to errors that are correlated with the fractional part of the disparity due to the so called \emph{pixel locking effect}~\cite{Shimizu2005, Miclea2015}. Fig.~\ref{fig:fract_part_correllation} illustrates the issue in more details; the mean error of the prediction is a function of the disparity fractional part. This, in turn, leads to systemic artifacts, \eg when the disparity is used to reconstruct a 3D view of a scene in stereo vision; see Section~\ref{sec:experimental:oned} for more details. The mitigation strategies proposed in the literature exploit the fact that the error caused by the pixel-locking effect is periodic, with a period of one pixel, and anti-symmetric over half that period~\cite{Shimizu2005}. This property can be exploited to average out the error. Since the disparity estimate for the (interpolated) image patch that is half a pixel away from the source patch should have an error that is approximately equal to minus the error of the reference patch,  the initial disparity estimate can be refined by combining it with the disparity estimate of the interpolated patch (plus or minus $0.5$ pixel depending in which direction this patch was interpolated). This method is efficient at removing the artifacts introduced by the pixel locking effect, but does not aim at improving the accuracy of the subpixel disparity. In practice, the method has a smoothing effect, which can also lead to some improvement. However, this smoothing effect also makes the method redundant with additional regularization that can be applied afterwards on the solution.

\vspace{6pt}
\noindent\textbf{Image space interpolation.} Image space interpolation methods lead to reconstruction errors that are decorrelated from the fractional part of the disparity.  Several papers, \eg~\cite{4409212, 6460147}, have demonstrated that these methods improve subpixel accuracy   since  they usually lead to better estimates of the matching cost for subpixel disparities. However, these methods use  exhaustive search over subpixel grids, which makes them computationally far less efficient compared to cost volume-based interpolation  methods and often unsuitable for real time applications~\cite{Cheng_2013}. Using local optimisation methods such as iterative gradient descent \cite{drones3030069} improves the computation time. All these methods can be made faster by using early-stopping criteria, but this will introduce quantization noise.

Different  strategies have been proposed to speed up image space interpolation methods. For example, Gehrig and Franke~\cite{4409212}  perform refinement only for  small disparities. This  makes sense for applications such as stereo vision where  small errors in pixel coordinates lead to large errors in depth estimation only for small disparities. Mizukami \etal~\cite{6460147}, on the other hand, use a coarse-to-fine approach where the finer disparity is computed only around each disparity. 

A point that has been overlooked by those methods is that, for most of the simple image interpolation methods used in practice, closed-form formulae for the refined disparity estimators can be derived. So far, this has only been considered in the case of one dimensional disparity for the Normalized Cross Correlation (NCC) cost function~\cite{1541350,Delon2007}. For the Sum of Square Differences (SSD) cost function, the formula is trivial and can be generalized to higher dimensions using a process known as predictive interpolation \cite{8960959, LMedSqFlow}; see Section~\ref{sec:formulae_higher_dims:feature_interp}. To the best of our knowledge, no prior work solved the case of the Sum of Absolute Differences (SAD) cost function. 

In this work, we propose and compare formulations for all three  common cost functions and generalize them to higher dimension patch-based matching. 

\vspace{6pt}
\noindent\textbf{Multidimensionnal subpixel refinement.} Multidimensional disparities received a little less attention than unidimensional disparities. The usual approach, when working in higher dimensions, is to assume that the cost function is \textbf{(1)} separable, \ie each dimension can be refined  independently of the others, and \textbf{(2)} isotropic, \ie  the separable property stays true  for any orientation of the pixel coordinate system~\cite{Nobach2005, debella2010sub}. Under these assumptions, any method that is applicable for unidimensional disparity generalizes to multidimensional disparities. This, however, adds a bias to the subpixel disparity estimate if the isotropy hypothesis is not respected, which is often the case in practice. Shimizu and Okutomi~\cite{2DSimultaneousSupixel} proposed a method to jointly estimate the horizontal and vertical subpixel disparities  for bi-dimensional area-based matching. The method assumes that the local subpixel optimums for each dimension, knowing all the other subpixel shifts, lie on a line. The intersection of the lines  for each dimension gives the optimal multidimensional subpixel alignment. While this method can be generalized to any subpixel interpolation method, it has only been evaluated for parabola fitting. It also requires to consider a large neighborhood around the optimal position; see Section~\ref{sec:formulae_higher_dims:cost_interp:anisotropic} for more details. Another option is to fit a multidimensional function, \eg a paraboloid, to the cost volume \cite{8960532}. This allows for more flexibility, but only applies to parabola-based cost interpolation as pyramid functions, which would generalize equiangular based refinement in one dimension, do not lead to simple closed form formulations in higher dimensions. 

\section{Notations and definitions}
\label{sec:notations}

We describe in this section the notation we use throughout the paper. In particular, we adopt a tensor notation so that all patch-based matching formulae can be written using basic linear algebra expressions and are easy to translate to the tensor processing libraries that are the standards in nowadays scientific computing.

\subsection{Tensor indexing and operations}

Tensors are denoted with bold uppercase symbols, \eg $\tensor{T}$. The lowercase letters subscripted with a number  $p_1,\cdots,p_n$  are used to index the spatial dimensions of the tensors (\eg vertical and horizontal dimensions in images), while $d_1,\cdots,d_n$ are used to index disparity dimensions, $c_1,\cdots,c_n$ to index a channel or a feature dimension, and $k_1,\cdots,k_n$ are used for arbitrary dimensions (\eg to refer to the other types of indices). A bold index letter without a number represents a full index, \eg $\coord{k}=\left(k_1,\cdots,k_n\right)$. Tensor indices are written in brackets $[\cdot]$ to distinguish them from other subscript notations, \eg the value of channel $c_1$ at pixel $p_1,p_2$ in the source image $\tensor{I}_s$ is denoted by $\tensor{I}_{s[p_1,p_2,c_1]}$. The absolute value $\lvert k_{i\tensor{T}} \rvert$ of a single index $k_i$ subscripted with a tensor $\tensor{T}$ is used to express the size of the corresponding dimension in $\tensor{T}$. $\Dim_{\tensor{T}}$ represents the number of dimensions of $\tensor{T}$. A dot $\cdot$ in place of an index indicates that all the elements on that axis are considered, \eg $\tensor{I}_{s[p_1,p_2,.]}$ represents all channels of the source image at pixel $p_1,p_2$, treated as a vector.

We write as $\patchingOp[l]{W}$ the patching operator with window $\tensor{W}$. For an image tensor $\tensor{I}$ it is defined as:
\begin{equation}
    \left(\patchingOp[l]{W} \tensor{I}\right)_{[p_1,p_2, c]} =  \tensor{I}_{[p_1+W_{[c,1]},p_2+W_{[c,2]},W_{[c,3]}]},
\end{equation}

\begin{figure}
    \centering
    \includegraphics[width=0.5\textwidth]{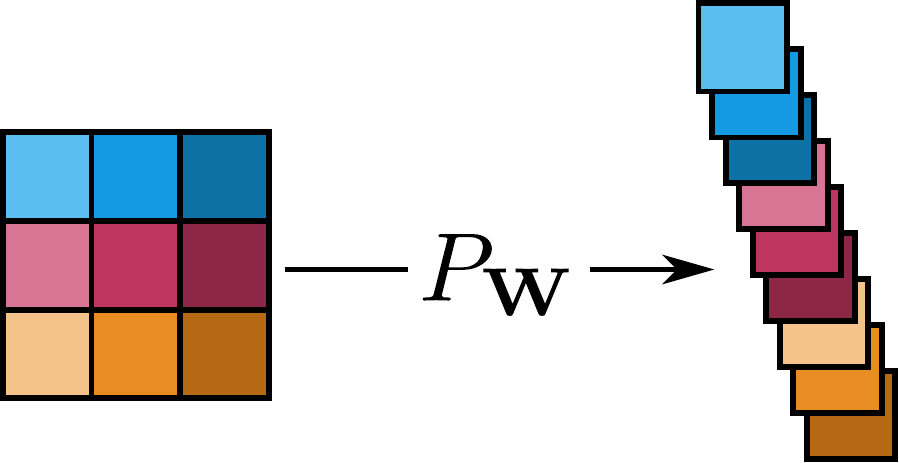}
    \caption{Visual representation of the patching operator, when used to extract a feature vector from an image patch.}
    \label{fig:unfold_op_visual}
\end{figure}

\noi This simplifies the problem of patch-based matching as search windows can be collapsed into a single pixel coordinate, all the information now being spread only across the channels, as shown in Figure~\ref{fig:unfold_op_visual}.

\subsection{Patch-based matching}

Patch-based matching is the process of finding for each patch in the source image $\tensor{I}_s$, the coordinates of the corresponding patch in the target image $\tensor{I}_t$. In what follows, $I$ without a subscript can refer to either the source or the target image. The patches are defined with a search window $\tensor{W}$, usually a square, centered at the reference pixel. In general, instead of working with the image pixels, features are used (\eg to speed up the matching process).  While many different, possibly non-linear operators, can be used to construct the features (\eg  grey-code filters kernels \cite{GreyCodeKernels}, SIFT \cite{790410}, or a full convolutional network for modern deep learning architectures \cite{Han_2015_CVPR}), in this work we focus on the patching operator to get the features for the evaluation of the proposed formulae. This way, a feature volume $\tensor{F}$ is given by:
\begin{equation}
    \tensor{F} = \patchingOp[l]{W}\tensor{I}, 
\end{equation}


\noindent with $l = 2$ for both grayscale and color images ensuring that $\tensor{F}$ is always a three dimensional tensor. This formulation allows us to express the cost volume $\tensor{C}$ by indexing the single dimension of the feature $\tensor{F}$ instead of multiple spatial dimensions of $\tensor{I}$. 


When invariance to intensity scale and constant bias is required, the feature volume needs to be whitened  by subtracting the mean and scaling it so that it has a unit variance. In what follows, $\tensor{ZF}$ refers to the zero-mean feature volume, $\tensor{NF}$ to the normalized feature volume, and $\tensor{ZNF}$ to the normalized zero-mean feature volume.

Various cost functions can then be computed from these feature volumes. Popular examples \cite{4270273} include the cross-correlation of the normalized features, called also Normalized Cross Correlation (NCC), the Sum of Square Differences (SSD),  and the Sum of Absolute Differences (SAD). Their zero mean counterparts, ZNCC, ZSSD, and ZSAD, are also popular choices. Those cost functions can be computed as:
\begin{equation}
    NCC_{[\pixel,\disparity]} = \left\langle \tensor{NF}_{s[\pixel,.]} , \tensor{NF}_{t[\pixel + \disparity,.]}\right\rangle,
\end{equation}

\begin{equation}
    SSD_{[\pixel,\disparity]} = \left\lVert \tensor{F}_{s[\pixel,.]} - \tensor{F}_{t[\pixel+\disparity,.]}\right\rVert_2^2,
\end{equation}

\begin{equation}
    SAD_{[\pixel,\disparity]} = \left\lVert \tensor{F}_{s[\pixel,.]} - \tensor{F}_{t[\pixel+\disparity,.]}\right\rVert_1.
\end{equation}

\noindent Where the $\disparity$ can be one-dimensional in the case of rectified stereo matching or bi-dimensional in the case of optical flow estimation or unrectified stereo matching. The zero-mean versions of the above similarity functions can be computed by using $\tensor{NF}$ instead of $\tensor{F}$, respectively $\tensor{ZNF}$ instead of $\tensor{NF}$. Some papers have proposed more exotic cost options like, for example, the Median of Squares (MedS) as more robust alternative for patch based matching \cite{LMedSqFlow}. The MedS offers a good robustness to noise but finding the optimal solution of the Least Median of Squares (LMedS)  is a non trivial combinatorial problem \cite{giloni2002least}. Existing methods addressed this issue using a variant of the RANSAC algorithm \cite{10.1145/358669.358692} to reduce the size of the search space. As such, they do not offer the speed and guarantee of the closed form solutions which can be derived for the NCC or SSD cost measures (and SAD in the one dimensional case). We included the MedS (and ZMedS) costs in our experimental comparison (Section \ref{sec:experimental}), but we did not include the mathematical derivations since there was no close form solution.

In the general case, the cost function can be expressed as:
\begin{equation}
    \tensor{C}_{[\pixel,\disparity]} = \costMeasure\left(\tensor{(Z)(N)F}_{s[\pixel,.]} , \tensor{(Z)(N)F}_{t[\pixel+\disparity,.]}\right),
\end{equation}

\noindent where $\costMeasure$ is a distance or similarity function, which takes as input two vectors from $\mathbb{R}^{\lvert c_{\tensor{(Z)(N)F}_{f}} \rvert}$.  The raw disparity $\nround{d}$ is  computed from the constructed cost volume and then refined, using some refinement formulae, to obtain a correction term $\Delta \hat{d}$ which can be added to the discrete disparity to get the final estimate of the true disparity: 
\begin{equation}
    \hat{d} = \nround{d} + \Delta \hat{d}.
\end{equation}

\begin{figure}
    \centering
    \includegraphics[width=0.7\textwidth]{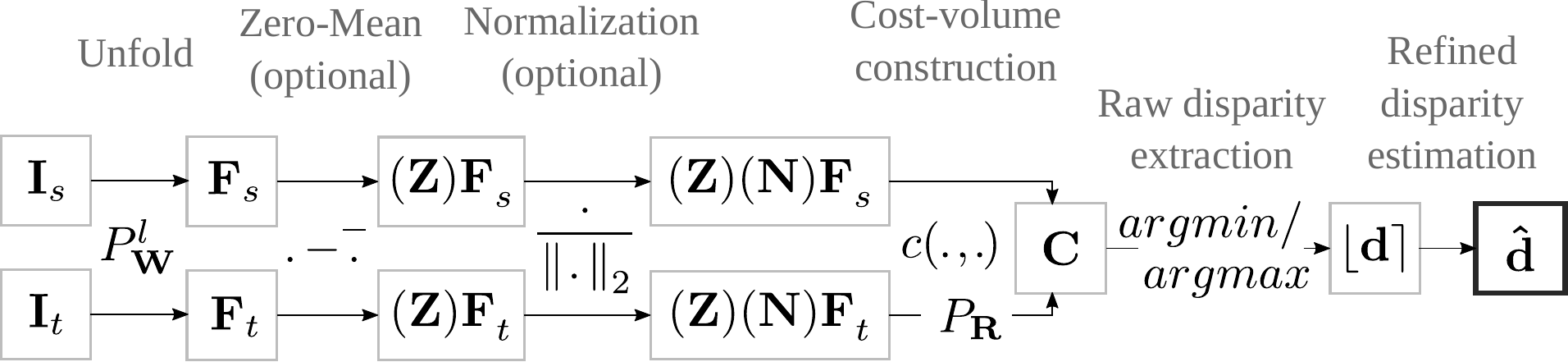}
    \caption{Generic pipeline for local patch-based matching.}
    \label{fig:patch_based_matching_pipeline}
\end{figure}

\noi Figure~\ref{fig:patch_based_matching_pipeline} summarizes the generic pipeline for patch-based matching. Alternatively, the process of finding discrete matches can be implemented using an algorithm like PatchMatch \cite{barnes2009patchmatch, CoherencySensitiveHashing} instead of an extensive search approach to exploit the image consistency and reduce the required number of cost computations.

\section{Subpixel interpolation formulae for unidimensional disparities}
\label{sec:interpFormulae}

There are two main approaches to obtain $\Delta \hat{d}$: \textbf{(1)} interpolating from the discrete cost volume values (Section~\ref{sec:costInterpFormulae}), or \textbf{(2)} interpolating the feature values to build a continuous cost volume  (Section~\ref{sec:imageInterpFormulae}). In this section, we review the formulae for both approaches and derive the corresponding closed form formulae to obtain the subpixel position in the case of uni-dimensional disparities.

\subsection{Cost volume-based interpolation}
\label{sec:costInterpFormulae}

The two most direct methods for cost volume-based interpolation are equiangular lines-based and parabola-based \cite{Shimizu2005}. For \textbf{equiangular lines-based interpolation}, $\Delta \hat{d}$ is the solution of the following equation:
\begin{equation}
    \zeta (\Delta \hat{d} + 1) + C_{[d-1]} = -\zeta (\Delta \hat{d} - 1) + C_{[d+1]},
\end{equation}

\noindent with:
\begin{equation}
    \zeta = \dfrac{C_{[d]} - C_{[d-1]}}{\lvert C_{[d]} - C_{[d-1]} \rvert} \max(\lvert C_{[d]} - C_{[d-1]} \rvert, \lvert C_{[d+1]} - C_{[d]} \rvert).
\end{equation}

\noindent Solving this equation gives:
\begin{equation}
\label{eq:equiangular_refinement}
    \Delta \hat{d} = \dfrac{C_{[d+1]} - C_{[d-1]}}{ 2 \zeta} .
\end{equation}

\noindent In the case of \textbf{parabola-based interpolation}, $\Delta \hat{d}$ is given by:
\begin{equation}
\label{eq:parabola_refinement}
    \Delta \hat{d} = \dfrac{C_{[d-1]} - C_{[d+1]}}{ 2(C_{[d+1]} - 2C_{[d]} + C_{[d-1]})}.
\end{equation}

\noi Alternatively, instead of a parabola, a Gaussian can be used to approximate the cost function \cite{Westerweel_1997}. This leads to a similar estimator as Equation~\eqref{eq:parabola_refinement}, but using the log of the cost instead of the cost itself. Note that this makes sense only if the cost or score function has been transformed to represent a probability distribution. Otherwise,  negative and zero values can lead to numerical singularities. As there is not a unique choice for such transformation, we will just ignore the Gaussian-based refinement in our comparison.

\subsection{Feature space-based interpolation}
\label{sec:imageInterpFormulae}

Instead of using the discrete cost or score values obtained during matching, one can generate a continuous cost volume by interpolating the features. In the case of unidimensional matching, when using linear interpolation, $\Delta \hat{d}$ can be computed using closed-form formulae for all the cost functions described in Section~\ref{sec:notations}. The continuous cost function is expressed as:
\begin{equation}
    C(\pixel,\disparity) = \costMeasure\left(\tensor{(N)F}_{s[\pixel,.]} , (N)F(\disparity)\right),
\end{equation}

\noindent where $F(\disparity) = (1 - \Delta d) \tensor{F}_{t[\pixel+\lfloor \disparity \rfloor],.} + \Delta d \tensor{F}_{t[\pixel+\lceil \disparity \rceil,.]}$ is the interpolated target feature vector and $\Delta d = \disparity - \lfloor \disparity \rfloor$ is the subpixel adjustment. 

Note that all the equations in this section are also valid when replacing $F$ by its zero-mean version $ZF$. In fact, this is true for feature volumes obtained by any linear convolution operator that can be used on the images to generate the features. For the normalized versions $NF$ and $ZNF$, the formulae depending on the derivative of the features with respect to the disparity need to be adapted. 

In the case of  normalized feature volumes,  the features can be interpolated either \textbf{(a)} prior to normalization, when $\tensor{F}_{s}$ and $\tensor{F}_{t}$ are proportional to the image patches' pixels intensities, or \textbf{(b)} after normalization, when $\tensor{NF}_{s}$ and $\tensor{NF}_{t}$ are used for interpolation, which is done on the unit sphere. For most applications, the norm of image patches is  highly spatially autocorrelated. This means that the difference between the norm of two successive patches will be similar. In this case, interpolating the features before or after normalizing them does not make a big difference.

The optimum subpixel correction $\Delta \hat{d}$ is reached when the derivative of the cost function with respect to $\Delta d$:
\begin{equation}
    \dfrac{d C}{d \Delta d} = \dotprod{\dfrac{\partial (N)F}{\partial \Delta d}}{\nabla_{F} \costMeasure},
\end{equation}

\noindent is equals to zero. Here, $\nabla_{F} \costMeasure$ is the gradient of the cost comparison function with respect to the (possibly normalized) features $(N)F$. $\dfrac{\partial (N)F}{\partial \Delta d}$ is the derivative of $(N)F$ with respect to the subpixel correction. For linear interpolation, the derivative of the features is constant in a given pixel interval:
\begin{equation}
\label{eq:unnormalizedfeaturesderivative}
    \dfrac{\partial F(\disparity)}{\partial \Delta d} = \left(\patchingOp{R}\tensor{F}_{t}\right)_{[\pixel,.,\lceil \disparity \rceil]} -  \left(\patchingOp{R}\tensor{F}_{t}\right)_{[\pixel,.,\lfloor \disparity \rfloor]}.
\end{equation}

\noindent For the normalized features, the expression becomes:
\begin{equation}
\label{eq:normalizedfeaturesderivative}
\dfrac{\partial NF}{\partial \Delta d} = \dfrac{\dfrac{\partial F(\disparity)}{\partial \Delta d}  \dotprod{F(\disparity)}{ F(\disparity)} - F(\disparity) \dotprod{F(\disparity)}{ \dfrac{\partial F(\disparity)}{\partial \Delta d}} }{ \dotprod{F(\disparity)}{ F(\disparity) }^{\dfrac{3}{2}} }.
\end{equation}

\noindent Note that, when setting the derivative of the cost function to $0$, the denominator can be ignored as it is bounded, meaning that the different expressions derived from Equation~\eqref{eq:normalizedfeaturesderivative} can usually be simplified.

\subsubsection{Normalized Cross Correlation (NCC)}
Let $\costMeasure$  be the cost function used to compute the cost volume $C$. When $\costMeasure$ is the  normalized cross correlation then
\begin{equation}
    \nabla_{F} \costMeasure = \tensor{F}_{s[\pixel,.]}.
\end{equation}

\noindent Using Equation~\eqref{eq:normalizedfeaturesderivative} to get the total derivative and setting it to zero,  the formula for the optimal subpixel correction in the unidimensional disparity refinement case becomes:
\begin{equation}
\label{eq:nccsubpixelunidimentional}
    \boxed{\Delta \hat{d}_{NCC} = \dfrac{a_{[0,1,0]} - a_{[1,0,0]}}{a_{[0,1,0]} -a_{[0,1,1]} - a_{[1,0,0]} + a_{[1,1,0]}},}
\end{equation}

\noindent where $a_{[\delta_0,\delta_1,\delta_2]}$ is defined as $\dotprod{ \tensor{F}_{s[\pixel,.]}}
{
\tensor{F}_{t[\pixel+\lfloor \disparity \rfloor + \delta_0,.]}
} $ times $\dotprod{\tensor{F}_{[\pixel+\lfloor \disparity \rfloor + \delta_1,.]}}{\tensor{F}_{t[\pixel+\lfloor \disparity \rfloor + \delta_2,.]}}$. 

While structured differently, this formula is strictly equivalent to the one presented by Psarakis et Evangelidis \cite{1541350} and  Delon et Roug{\'e} \cite{Delon2007}. Psarakis et Evangelidis \cite{1541350} and  Delon et Roug{\'e} \cite{Delon2007} have overlooked the important fact that using a (Z)NCC-based formula for subpixel refinement can mitigate some of the biases inherent in linear interpolation in addition to the benefits of using normalized cross-correlation (i.e., being insensitive to differences of gain or bias between the source and target images). A detailed explanation of why this is the case can be found in Section~II in the Supplementary Material.

\subsubsection{Sum of Square Differences (SDD)}

When the cost function used to compute the cost volume is the sum of square differences, then:
\begin{equation}
    \nabla_{F} \costMeasure = 2F(\disparity) - 2\tensor{F}_{s[\pixel,.]}.
\end{equation}

\noindent Since the sum of square differences does not rely on normalized features, Equation~\eqref{eq:unnormalizedfeaturesderivative} gives the derivative of the features. This means that the optimal subpixel correction $\Delta \hat{d}_{SSD}$ can be computed as:
\begin{equation}
\label{eq:ssdsubpixelunidimentional}
   \boxed{\Delta \hat{d}_{SSD} = \dfrac{\dotprod{\dfrac{\partial F(\disparity)}{\partial \Delta d}}{ \tensor{F}_{s[\pixel,.]} }
}{\dotprod{\dfrac{\partial F(\disparity)}{\partial \Delta d}}{\dfrac{\partial F(\disparity)}{\partial \Delta d}}} .}
\end{equation}

\noindent The normalized variants of SSD, NSSD and ZNSSD are strictly equivalent to NCC, respectively ZNCC \cite{ncc_nssd_equiv}. This means that those variants can be treated with Equation~\eqref{eq:nccsubpixelunidimentional}.

\subsubsection{Sum of Absolute Differences (SAD)}
\label{sec:sadweigthedmedian}

When the sum of absolute differences is used as the matching function, the gradient of the cost becomes: 
\begin{equation}
    \nabla_{F} \costMeasure = \sgn \left(F(\disparity) - \tensor{F}_{s[\pixel,.]}\right),
\end{equation}

\noindent which is not a regular function. However, the cost function $C_{SAD}$ is still continuous and convex between two integer pixel disparities when the features are linearly interpolated in the interval. Its derivative is a piecewise constant function given by: 
\begin{equation}
    \dfrac{d C_{SAD}}{d \Delta d} = \dotprod{ \dfrac{\partial F(\disparity)}{\partial \Delta d} }{\sgn \left(F(\disparity) - \tensor{F}_{s[\pixel,.]}\right)}.
\end{equation}

\noindent The optimal subpixel correction $\Delta \hat{d}_{SAD}$ is the only point for which this derivative would be positive on one side and negative on the other. This point can be computed with a weighted median \cite{Sabo2011}:
\begin{equation}
    \boxed{\Delta \hat{d}_{SAD} = \median_{\tensor{\omega}} \Delta \tensor{\mathring{d}} ,}
\end{equation}

\noindent where the potential subpixel corrections $\Delta \tensor{\mathring{d}}$ are given by:
\begin{equation}
    \Delta \tensor{\mathring{d}}_{[c]} = \dfrac{\tensor{F}_{s[\pixel,c]} - \tensor{F}_{t[\pixel,c,\lfloor \disparity \rfloor]}}{\tensor{F}_{t[\pixel+\lceil \disparity \rceil,c]} - \tensor{F}_{t[\pixel+\lfloor \disparity \rfloor,c]}}.
\end{equation}
\noindent The weights $\tensor{\omega} = \lvert \dfrac{\partial F(\disparity)}{\partial \Delta d} \rvert$. Given $\Delta \tensor{\mathring{d}}$ and the corresponding weights $\tensor{\omega}$, we have:
\begin{equation}
    \Delta \hat{d}_{SAD} = \Delta \tensor{\mathring{d}}_{[\hat{c}]},
\end{equation}

\noindent with 
\begin{equation}
    \hat{c} = \argmax_{c \in \Naturals} 2\sum_{i=1}^{c} \tensor{\omega}_{[i]} \le \sum_{i=1}^{\lvert c_{\tensor{\omega}} \rvert} \tensor{\omega}_{[i]}.
\end{equation}

\noi This can effectively be solved in $O(\lvert c_{\tensor{F}} \rvert)$ time using a median selection algorithm, which is asymptotically not worse than the complexity of the other cost functions. 

\section{Subpixel interpolation formulae for multidimensional disparities}
\label{sec:formulae_higher_dims}

The subpixel refinement formulae presented in the previous section are limited to one-dimensional patch-based matching, \eg rectified stereo images. 
However,  we are also interested in formulae for patch-based matching with multidimensional disparities. The multidimensional case is more general and has a wide range of applications, \eg optical flow estimation. In what follows, Section~\ref{sec:formulae_higher_dims:cost_interp} covers the cost volume interpolation-based formulae while Section~\ref{sec:formulae_higher_dims:feature_interp} introduces the generalized feature volume-based formulae.

\subsection{Using cost volume interpolation}
\label{sec:formulae_higher_dims:cost_interp}

Traditional cost volume-based closed-form formulae for subpixel refinement can be roughly divided into two categories: those that assume that the cost function is locally well approximated by an isotropic peak and those that assume the cost function is locally well approximated by a possibly anisotropic peak.

\subsubsection{Isotropic interpolation}
\label{sec:formulae_higher_dims:cost_interp:isotropic}

If the peak of the cost function is isotropic then the problem of subpixel refinement becomes separable and can be carried out independently on each dimension using any of the formulae presented in Section~\ref{sec:costInterpFormulae}~\cite{Nobach2005, debella2010sub}:
\begin{equation}
\label{eq:multidimseparable}
    \Delta \vec{\hat{d}} = \left(\begin{array}{c}
        \Delta \hat{d}(C_{[d_0-1,\cdots,d_n]}, C_{[d_0,\cdots,d_n]}, C_{[d_0+1,\cdots,d_n]})  \\
        \vdots \\
        \Delta \hat{d}(C_{[d_0,\cdots,d_n-1]}, C_{[d_0,\cdots,dn]}, C_{[d_0,\cdots,d_n+1]})
    \end{array}\right).
\end{equation}

\noindent This approach has many benefits. In particular, it allows for basically any closed-form formula to be generalized to arbitrarily high dimensions. However, this baseline hypothesis, \ie the peak of the cost function is isotropic, is not always valid, which can lead to an important bias in certain configurations \cite{2DSimultaneousSupixel}.

\subsubsection{Anisotropic interpolation}
\label{sec:formulae_higher_dims:cost_interp:anisotropic}

\begin{figure}
    \centering
    \includegraphics[width=0.5\textwidth]{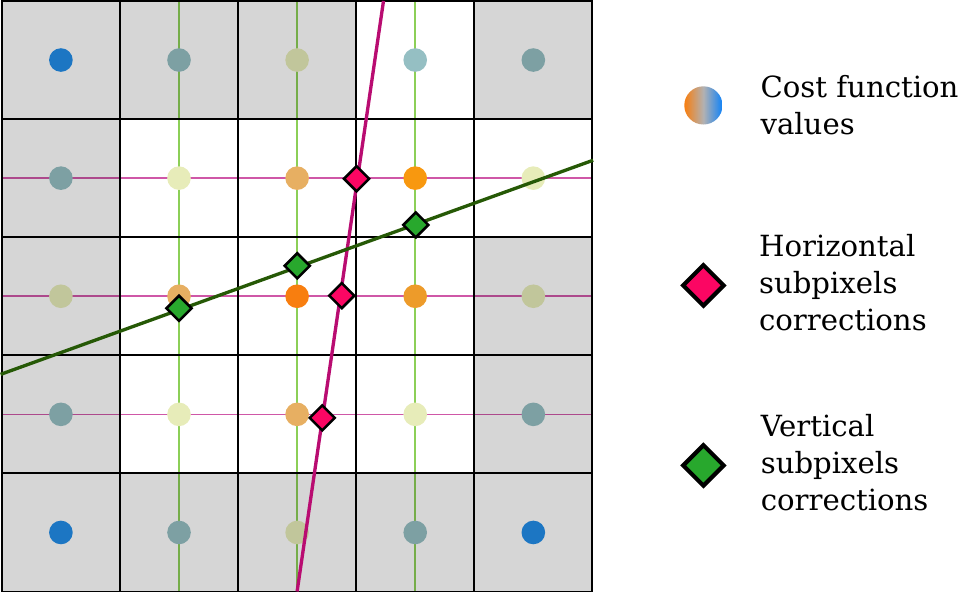}
    \caption{Visual representation for the anisotropic cost interpolation. Pixels required for the computation are in white. The optimum is at the intersection of both lines fitting the subpixels corrections.}
    \label{fig:2d_cost_anisotropic}
\end{figure}

To account for possible anisotropy in the cost function peak, Shimizu and Okutomi~\cite{2DSimultaneousSupixel} proposed a procedure where a line is fitted to the subpixel correction in one direction keeping all the other directions as constant. The optimal two-dimensional subpixel disparity then lies at the intersection of all the lines. Figure~\ref{fig:2d_cost_anisotropic} gives a visual representation of the approach.

This approach has the same benefits as the separable approach presented in Section~\ref{sec:formulae_higher_dims:cost_interp:isotropic}, \ie it allows for basically any closed form formula to be generalized to arbitrarily high dimensions. The approach, however, has some limitations. For instance, the optimal discrete pixel adjustment for a given dimension can be different from zero if the index in the other dimension is different from zero. It means that the method might require values in the cost function arbitrarily far away from the discrete optimum. It also means that both the vertical and horizontal subpixel correction lines can be very close to being co-linear, which  can lead to numerical instability. Shimizu and Okutomi~\cite{2DSimultaneousSupixel} recommend to only consider a radius of two pixels around the discrete optimum and fall back to a default method beyond this point. Although the method can be generalized to more than two dimensions, it can quickly become impractical  beyond two dimensions as the number of subpixel positions to compute grows geometrically with the number of dimensions of the disparities.

An alternative solution is to fit a multidimensional function (\eg a paraboloid) to the data \cite{8960532}. This approach seems more natural but is difficult to generalize to any formulation (\eg for equiangular refinement).

\subsection{Generalized feature volume-based refinement with barycentric and predictive interpolation}
\label{sec:formulae_higher_dims:feature_interp}

%

The approaches presented in Section~\ref{sec:imageInterpFormulae} can be reinterpreted as solving a multidimensional optimization problem that finds   $\vec{\hat{\beta}}$ such that:
\begin{equation}
    \label{eq:barycentric_refinement_problem}
    \vec{\hat{\beta}} = \argmin_{\vec{\beta}} \costMeasure(\matr{A}\vec{\beta}, \vec{f}_{s}),
\end{equation}

\noindent with the additional constraints that $\vec{\beta}$ represents a set of barycentric coordinates, \ie $\sum_{\beta_i \in \vec{\beta} } \beta_i = 1$. $\vec{f}_{s}$ is a reference feature vector from the source feature volume and $A = \left[ \vec{f}_{t,1},...,\vec{f}_{t,n} \right]$ is a matrix whose columns are the feature vectors from the target feature volume between which the interpolation is performed. 

This approach can be generalized to higher dimensions as long as a closed-form formula to solve Equation~\eqref{eq:barycentric_refinement_problem} exists when $\text{dim}(\vec{\beta}) > 2$. In this section, we give such closed-form formula for the (Z)NCC and (Z)SSD matching functions. Unfortunately, there is no closed-form formula for the (Z)SAD matching function \cite{SOVICKRZIC2018119}. However, we propose an algorithm that is guaranteed to yield the correct solution in a finite number of steps.

\begin{figure*}[t]
    \centering
    \begin{subfigure}[t]{0.17\textwidth}
        \includegraphics[width=\textwidth]{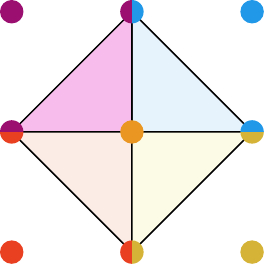}
        \caption{Features split with Rook contiguity}
        \label{fig:interp_2d:cut_rook}
    \end{subfigure}
    \hfill
    \begin{subfigure}[t]{0.17\textwidth}
        \includegraphics[width=\textwidth]{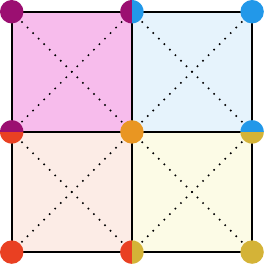}
        \caption{Features split with Queen contiguity}
        \label{fig:interp_2d:cut_queen}
    \end{subfigure}
    \hfill
    \begin{subfigure}[t]{0.17\textwidth}
        \includegraphics[width=\textwidth]{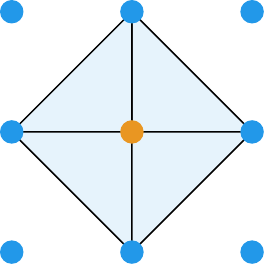}
        \caption{Features symmetric with Rook contiguity}
        \label{fig:interp_2d:sym_rook}
    \end{subfigure}
    \hfill
    \begin{subfigure}[t]{0.17\textwidth}
        \includegraphics[width=\textwidth]{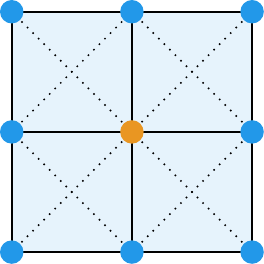}
        \caption{Features symmetric with Queen contiguity}
        \label{fig:interp_2d:sym_queen}
    \end{subfigure}
    
    \caption{The four considered approaches for 2D patch-based matching. Features are either split between the corners of the discrete match, or considered all at once. The considered features sets can be selected with either queen or rook contiguity.}
    \label{fig:interp_2d}
\end{figure*}

One constraint of barycentric interpolation is that it yields a unique interpolated value for each point if and only if the interpolation domain is isomorphic to a simplex (\ie a segment in 1D, a triangle in 2D, a tetrahedron in 3D, etc.). This is, however, not the case for a pixel (for 2D refinement) nor a voxel (for 3D refinement), which have four, respectively eight, corners. Two mitigation strategies can be used; see Figure~\ref{fig:interp_2d}: \textbf{(\subref{fig:interp_2d:cut_rook})} Consider only the coordinates within a given pixel or voxel for which the closest corner is the one pointed to by $\nround{\coord{\hat{d}}}$. Since this set is always isomorphic to a simplex, barycentric interpolation can be used. Another benefit is that the number of considered feature vectors grows linearly instead of geometrically with the number of considered dimensions. 

Alternatively, \textbf{(\subref{fig:interp_2d:cut_queen})} the problem can be relaxed so that the optimal weights $\vec{\beta}$ are selected even if multiple weight sets lead to the same refined disparity $\vec{\hat{d}}$. This approach is known as predictive interpolation \cite{8960959} where the interpolation weight of different image pixels is extracted from the patches themselves.
This can be interpreted as interpolating the disparity in the feature space instead of interpolating the features in the disparity space. While the continuous source or target feature volumes form complex and maybe even self-intersecting hypersurfaces, it is reasonable to assume they are locally smooth. It means that, functions (\eg the disparity) defined for each feature vector can be interpolated on said hypersurfaces. 

Predictive interpolation allows to treat problems with arbitrary shapes like the four corners of a pixel, but also to generalize the approach to multiple feature vectors in lower dimensions (\eg instead of treating the left and right pixels of $\nround{\coord{d}}$ with the equations presented in Section~\ref{sec:imageInterpFormulae}, both the left and the right pixels can be considered at the same time).

Another issue with barycentric interpolation is how to enforce the constraint on $\vec{\beta}$. The easiest way to do it is by performing a change of variable from the barycentric coordinates $\vec{\beta}$ to a set of affine coordinates $\vec{\alpha}$:
\begin{equation}
    \matr{M} \vec{\alpha} + \vec{f}_{t,n} = \matr{A}\vec{\beta},
\end{equation}

\noindent with $\matr{M} = \left[ \vec{f}_{t,1} - \vec{f}_{t,n},...,\vec{f}_{t,n-1} - \vec{f}_{t,n} \right]$ and $\alpha_i = \beta_i \forall i \in [1,n-1]$.

The final disparity $\coord{\hat{d}}$ is computed as:
\begin{equation}
    \coord{\hat{d}} = \matr{D}\vec{\hat{\beta}},
\end{equation}

\noindent where $\matr{D} = \left[\coord{\nround{d}}_{\vec{f_{t,1}}},...,\coord{\nround{d}}_{\vec{f_{t,n}}} \right]$ is the matrix formed with the discrete disparity coordinates of the considered target features vectors.

\subsubsection{Normalized Cross Correlation (NCC)} 

For normalized cross correlation, when more than two feature vectors are considered for interpolation, Equation~\eqref{eq:normalizedfeaturesderivative} does not simplify to a linear function of the barycentric coefficients. However, one still can get a closed-form formula for $\vec{\hat{\beta}}$ by approaching the optimization problem using geometry instead of calculus. 

\begin{figure}
    \centering
    \includegraphics[width=0.4\textwidth]{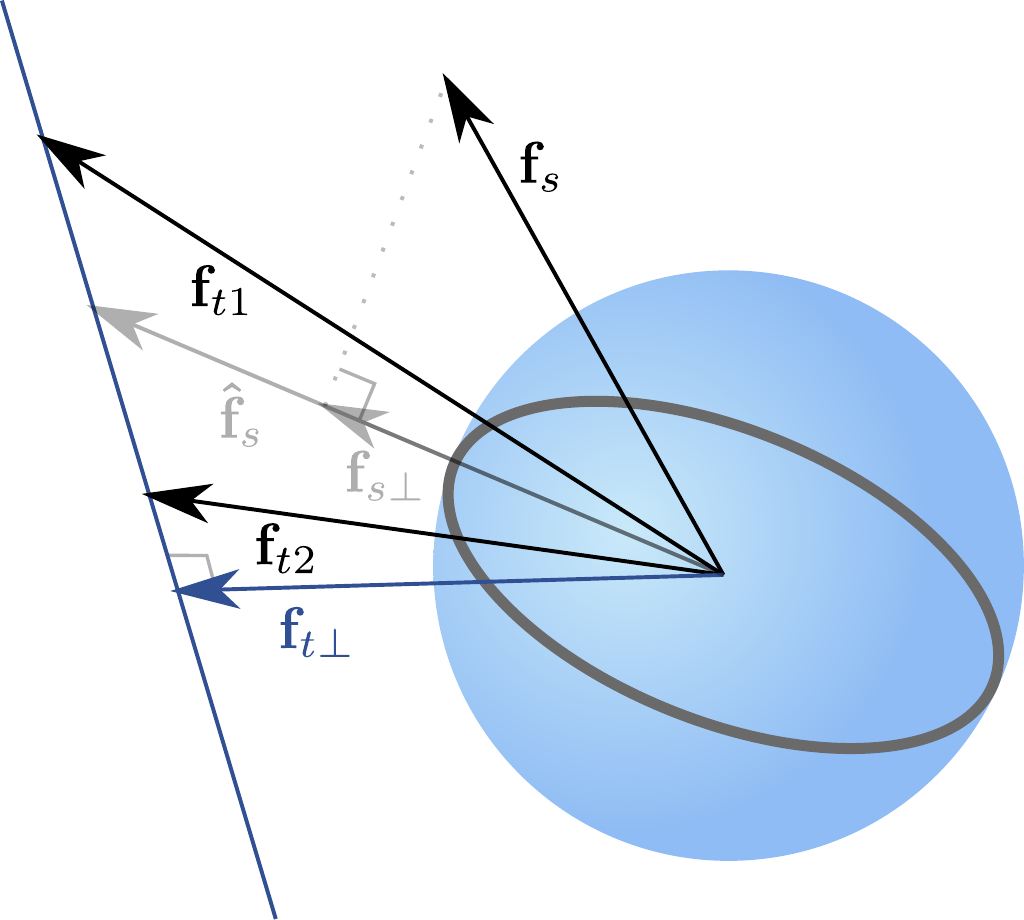}
    \caption{Visualizing the problem of barycentric coordinate regression in the feature vector space for the NCC cost function.}
    \label{fig:ncc_geometric_approach}
\end{figure}

The optimally interpolated feature vector $\Vec{\hat{f}}_{s}$ is the only vector in the space $\left\{ \matr{M}\vec{\alpha} + \vec{f}_{t,n} \mid \vec{\alpha} \in \Reals^{n-1} \right\}$ to be co-linear with the projection $\Vec{f}_{s\perp}$ of $\vec{f}_{s}$ onto $\spn \left( \matr{A} \right)$. This is because the normalized cross correlation is a dot product between feature vectors scaled to lie on the unit sphere. Thus, only the direction matters. Figure~\ref{fig:ncc_geometric_approach} presents visually the approach  for two target feature vectors in a 3D feature space.

The procedure to compute the affine coordinates $\Vec{\hat{\alpha}}$ of $\vec{\hat{f}}_{s}$ is as follow; \textbf{First}, compute $\Vec{f}_{s\perp}$, which is given by:
\begin{equation}
    \Vec{f}_{s\perp} = \matr{A} \left( \matr{A}^\top \matr{A} \right)^{-1} \matr{A}^\top \Vec{f}_{s}.
\end{equation}

\noindent \textbf{Second}, find the vector $\Vec{f}_{t\perp}$, which is the closest vector in $\left\{  \matr{M}\vec{\alpha} + \vec{f}_{t,n} \mid \vec{\alpha} \in \Reals^{n-1} \right\}$ to the origin:
\begin{equation}
    \Vec{f}_{t\perp} = \vec{f}_{t,n} - \matr{M} \left( \matr{M}^\top \matr{M} \right)^{-1} \matr{M}^\top \left( \vec{f}_{t,n} \right).
\end{equation}

\noindent Now, $\Vec{\hat{f}}_{s}$ can be computed as follow:
\begin{equation}
    \Vec{\hat{f}}_{s} = \dfrac{\dotprod{\Vec{f}_{t\perp}}{\Vec{f}_{t\perp}}}{\dotprod{\Vec{f}_{t\perp}}{\Vec{f}_{s\perp}}} \Vec{f}_{s\perp}.
\end{equation}

\noindent Finally, $\Vec{\hat{\alpha}}$ can be computed as:
\begin{equation}
    \boxed{\Vec{\hat{\alpha}} = \left( \matr{M}^\top \matr{M} \right)^{-1} \matr{M}^\top \left( \Vec{\hat{f}}_{s} - \vec{f}_{t,n} \right).}
\end{equation}

\noindent Note that using least squares is not required here since  we know that $\Vec{\hat{f}}_{s} \in \left\{ \matr{M}\vec{\alpha} + \vec{f}_{t,n} \mid \vec{\alpha} \in \Reals^{n-1} \right\}$. However, it is convenient since either $\left( \matr{M}^\top \matr{M} \right)^{-1} \matr{M}^\top $ or the QR factorization of $\matr{M}$ has already been computed to get $\Vec{f}_{t\perp}$. 

This approach is equivalent to Equation~\eqref{eq:nccsubpixelunidimentional} when using only two target feature vectors. The proof, which we generated using the computer algebra system Maxima \cite{maxima}, is provided in the Supplementary Material.

\subsubsection{Sum of Square Differences (SAD)}
\label{sec:generalizedsad}

For the sum of square differences, the formula given in Equation~\eqref{eq:ssdsubpixelunidimentional} generalizes directly for any set of vectors used for the barycentric composition. We just have to replace $\dfrac{\partial F(\disparity)}{\partial \Delta d}$ by $\matr{M}$ to get $\Vec{\hat{\alpha}}$, and obtain:
\begin{equation}
     \boxed{\Vec{\hat{\alpha}} = \left( \matr{M}^\top \matr{M} \right)^{-1} \matr{M}^\top \left( \Vec{f}_{s} - \vec{f}_{t,n} \right).}
\end{equation}

\noindent This formula is equivalent to the one developed in~\cite{8960959} and \cite{LMedSqFlow} for predictive interpolation.

\subsubsection{Sum of Absolute Differences}
\label{sec:sadmultidimalg}

In the case of a barycentric composition with more than two target vectors, finding $\vec{\hat{\alpha}}$ is a least absolute difference problem of the form:
\begin{equation}
    \boxed{\vec{\hat{\alpha}} = \argmin_{\vec{\alpha}} \left\lVert (\Vec{f}_{s} - \vec{f}_{t,n}) - \matr{M} \vec{\alpha} \right\rVert_1 .}
\end{equation}

\noindent This problem can be transformed to a linear program. Thus, one can solve it using a variant of the simplex algorithm \cite{boyd2004convex}. However, this is unnecessarily complicated and suboptimal in practice, as this general method ignores much of the specificities of the least absolute difference problem. Instead, one can use gradient descent methods, which are guaranteed to reach the exact solution since the problem is convex. Dedicated variants of the algorithm, see for example~\cite{SOVICKRZIC2018119}, are guaranteed to reach the exact solution in a finite number of steps. 

The gradient of the cost with respect to $\vec{\alpha}$ is given by:
\begin{equation}
    \nabla_{\vec{\alpha}} \costMeasure = \matr{M}^\top \nabla_{\vec{f}} \costMeasure,
\end{equation}

\noindent with $\nabla_{\vec{f}} \costMeasure$ given by:
\begin{equation}
    \nabla_{\vec{f}} \costMeasure = \sgn \left( (\Vec{f}_{s} - \vec{f}_{t,n}) - \matr{M} \vec{\alpha} \right).
\end{equation}

\noi Starting from an initial guess $\vec{\mathring{\alpha}}_0$, the problem can be reduced to a uni-dimensional $\ell_1$ minimization along the direction of the gradient. Starting from the new estimate, the procedure iterates until it reaches $\vec{\hat{\alpha}}$. When the solution lies on the inflexion hyperplane of one of the absolute value components of the cost, some artifacts can occur since the gradient cannot be computed by a linear composition of partial derivatives. Each intermediate line-restricted optimum $\vec{\mathring{\alpha}}_i$ will be at the intersection of such hyperplanes. The solution is to constrain the optimization to the hyperplanes encountered in the previous optimization until the restricted space is uni-dimensional. When this occurs, the newly intersected hyperplane has to be added to the constraints and one previous hyperplane has to be deselected. The left and right gradients can be computed for each selected hyperplane. The hyperplane leading to the steepest descent when unselected is the one to remove from the constraints. As the method will end up following the edges of the cost function and will always lead to a smaller value, it is guaranteed to converge in a finite time.

    
The perpendicular vectors of the hyperplanes currently selected as constraints are the corresponding rows of the matrix $\matr{M}$ and can be stored in a matrix $\matr{Z}$. The direction of the projected gradient is then given by:
\begin{equation}
    \nabla_{\vec{\alpha},\matr{Z}} \costMeasure = \nabla_{\vec{\alpha}}\costMeasure - \matr{Z}^\top(\matr{Z} \matr{Z}^\top)^{-1} \matr{Z} \nabla_{\vec{\alpha}}\costMeasure.
\end{equation}

\noindent Note that once a sufficient number of constraints are selected such that the space they span is unidimensional, the gradient is computed not to decide in which direction to optimize (this is given by the constraints) but to determine which constraint needs to be deselected. It also gives the direction $\vec{d}$ of the parametric line $\alpha(t) = \vec{d}t$ along which the optimization is performed.

To compute the optimal $\hat{t}_i$ given the optimization direction $\vec{d}_i$, the weighed median formula presented in Section~\ref{sec:sadweigthedmedian} can be used. The potential scaling parameters $\Delta \tensor{\mathring{t}}_i$ are given by:
\begin{equation}
    \Delta \tensor{\mathring{t}}_{i,[c]} = \dfrac{\vec{f}_{s,[c]} - \vec{f}_{t,n,[c]} - \matr{M} \vec{\alpha}_{i,[c]} }{ \matr{M} \vec{d}_{i,[c]} }
\end{equation}

\noindent and the weights $\tensor{\omega}_i$ are equal to $\lvert \matr{M}\vec{d}_{i,[c]} \rvert$. This leads to the following iterative formula for the $\vec{\mathring{\alpha}}_i$:
\begin{equation}
    \vec{\mathring{\alpha}}_{i+1} = \vec{\mathring{\alpha}}_{i} + \vec{d}_i \left( \median_{\tensor{\omega}_i} \Delta \tensor{\mathring{t}}_{i,[c]} \right).
\end{equation}

\section{Algorithmic complexity of the proposed formulae}
\label{sec:algo_complexity}

Discrete dense patch-based matching has a running complexity of $\bigO{S \times R \times F}$, with $S$ the size of the images, $R$ the size of the search range, and $F$ the length of the feature vectors. While $R$ usually represents a full dense search space, specific methods such as PatchMatch \cite{barnes2009patchmatch}, and its potential improvements like Coherency Sensitive Hashing \cite{CoherencySensitiveHashing}, can be used to reduce its size. However, in the best case, the running time complexity of dense patch-base matching would still be $\bigO{S \times F}$. For comparison, cost-based interpolation formulae have a running time complexity, when processing a whole image, of $\bigO{S}$. Image-based interpolation formulae, on the other hand, have a running time complexity of $\bigO{S \times F}$, which is higher than cost-based refinement methods. However, the overall complexity is still dominated by the dense matching step which has a complexity of $\bigO{S \times R \times F}$. Practically speaking, the difference in execution time between cost-based and image-based refinements is negligible when compared to the duration needed to establish correspondences in all but the more time critical applications. The algorithm proposed in Section~\ref{sec:generalizedsad}, on the other hand, is a variant of the simplex method and as such has a worst time complexity of $\bigO{S \times 2^F}$, which would increase the overall complexity of the dense patch-base matching problem. However, in practice, this is hardly an issue since, on average, the simplex still has a polynomial time complexity \cite{10.1145/990308.990310}. Also, since our method also works as a gradient descent method, it can be restarted for early iterations to speed up convergence by not following the edges of the function.

\section{Experimental results}
\label{sec:experimental}

In this section, we analyze the subpixel accuracy and resistance to the pixel locking effect of the proposed closed-form formulae for subpixel refinement. We first consider the case of rectified stereo pair matching (Section~\ref{sec:experimental:oned}), which is a  one-dimensional patch-based problem. Then, in Section~\ref{sec:experimental:twod}, we consider the problem of optical flow estimation, which is a two-dimensional patch-based matching problem. All results are provided when the interpolation of the feature volumes is performed before feature whitening. The Supplementary Material provides additional results including the case when the interpolation is performed after feature whitening. The datasets analyzed during the current study are available in the Middlebury vision repository, \url{https://vision.middlebury.edu/}.

\subsection{One dimensional patch-based matching}
\label{sec:experimental:oned}


\begin{table*}[t]
\centering
\begin{adjustbox}{max width=\textwidth}
\begin{tabular}{ll|rrrrrr|rrrrr|}
\cline{3-13}
 &  & \multicolumn{6}{c|}{Mean MAE over Middlebury 2014 {[}px{]}} & \multicolumn{5}{c|}{Mean SNR over Middlebury 2014 {[}dB{]}} \\ \hline
\multicolumn{1}{|l|}{$C$} & $W$ & \multicolumn{1}{c}{Raw} & \multicolumn{1}{c}{Parabola} & \multicolumn{1}{c}{Equiangular} & \multicolumn{1}{c}{\cite{Shimizu2005}} & \multicolumn{1}{c}{Barycentric} & \multicolumn{1}{c|}{Predictive} & \multicolumn{1}{c}{Parabola} & \multicolumn{1}{c}{Equiangular} & \multicolumn{1}{c}{\cite{Shimizu2005}} & \multicolumn{1}{c}{Barycentric} & \multicolumn{1}{c|}{Predictive} \\ \hline
\multicolumn{1}{|l|}{} & 5x5 & 0.26 & 0.15 & 0.15 & 0.15 & \textbf{0.13} & 0.16 & -14.16 & \textbf{-27.80} & -22.08 & -23.16 & -15.41\\
\multicolumn{1}{|l|}{} & 7x7 & 0.28 & 0.15 & 0.15 & 0.15 & \textbf{0.14} & 0.15 & -16.06 & -24.75 & -24.33 & \textbf{-25.34} & -16.31\\
\multicolumn{1}{|l|}{} & 9x9 & 0.28 & 0.15 & 0.15 & 0.15 & \textbf{0.15} & 0.16 & -18.07 & -23.40 & \textbf{-25.47} & -23.42 & -17.71\\
\multicolumn{1}{|l|}{\multirow{-4}{*}{NCC}} & 11x11 & 0.29 & 0.16 & 0.16 & \textbf{0.16} & 0.16 & 0.17 & -20.14 & -22.66 & \textbf{-26.30} & -22.19 & -19.00\\\hline
\multicolumn{1}{|l|}{} & 5x5 & 0.26 & 0.15 & 0.15 & 0.14 & {\color[HTML]{006600}\textbf{0.12}} & 0.15 & -13.14 & \textbf{-26.68} & -20.70 & -26.12 & -15.14\\
\multicolumn{1}{|l|}{} & 7x7 & 0.27 & 0.15 & 0.15 & 0.14 & \textbf{0.13} & 0.15 & -15.01 & -25.57 & -23.32 & \textbf{-26.21} & -16.06\\
\multicolumn{1}{|l|}{} & 9x9 & 0.28 & 0.15 & 0.15 & 0.14 & \textbf{0.14} & 0.15 & -17.15 & -23.64 & \textbf{-25.08} & -24.23 & -17.49\\
\multicolumn{1}{|l|}{\multirow{-4}{*}{ZNCC}} & 11x11 & 0.29 & 0.16 & 0.16 & \textbf{0.15} & 0.15 & 0.16 & -19.41 & -22.79 & \textbf{-25.81} & -22.59 & -18.67\\\hline
\multicolumn{1}{|l|}{} & 5x5 & 0.28 & 0.18 & 0.18 & 0.18 & \textbf{0.17} & 0.20 & -18.77 & \textbf{-26.17} & -24.43 & -23.67 & -15.31\\
\multicolumn{1}{|l|}{} & 7x7 & 0.29 & 0.18 & 0.18 & 0.18 & \textbf{0.17} & 0.19 & -19.42 & \textbf{-25.30} & -24.99 & -24.79 & -16.41\\
\multicolumn{1}{|l|}{} & 9x9 & 0.29 & 0.18 & 0.18 & 0.18 & \textbf{0.17} & 0.19 & -20.75 & -23.79 & \textbf{-25.04} & -23.98 & -17.31\\
\multicolumn{1}{|l|}{\multirow{-4}{*}{SSD}} & 11x11 & 0.30 & \textbf{0.18} & 0.18 & 0.19 & 0.18 & 0.20 & -21.97 & -22.36 & \textbf{-24.56} & -22.37 & -17.93\\\hline
\multicolumn{1}{|l|}{} & 5x5 & 0.26 & 0.15 & 0.15 & 0.18 & \textbf{0.13} & 0.15 & -13.99 & \textbf{-27.88} & -18.06 & -24.49 & -16.15\\
\multicolumn{1}{|l|}{} & 7x7 & 0.28 & 0.15 & 0.15 & 0.18 & \textbf{0.14} & 0.15 & -16.01 & -24.71 & -20.06 & \textbf{-25.98} & -17.13\\
\multicolumn{1}{|l|}{} & 9x9 & 0.28 & 0.15 & 0.15 & 0.18 & \textbf{0.15} & 0.16 & -18.13 & -23.38 & -22.08 & \textbf{-23.46} & -18.42\\
\multicolumn{1}{|l|}{\multirow{-4}{*}{ZSSD}} & 11x11 & 0.29 & 0.16 & \textbf{0.16} & 0.18 & 0.16 & 0.16 & -20.11 & -22.59 & \textbf{-23.91} & -21.91 & -19.85\\\hline
\multicolumn{1}{|l|}{} & 5x5 & 0.29 & 0.19 & \textbf{0.18} & 0.18 & 0.19 & 0.23 & -16.29 & -22.19 & \textbf{-23.34} & -22.98 & -17.41\\
\multicolumn{1}{|l|}{} & 7x7 & 0.29 & 0.18 & \textbf{0.17} & 0.18 & 0.19 & 0.22 & -16.77 & -23.53 & \textbf{-23.77} & -23.70 & -18.52\\
\multicolumn{1}{|l|}{} & 9x9 & 0.29 & 0.18 & \textbf{0.17} & 0.18 & 0.20 & 0.22 & -17.97 & \textbf{-24.74} & -23.99 & -24.67 & -19.08\\
\multicolumn{1}{|l|}{\multirow{-4}{*}{SAD}} & 11x11 & 0.30 & 0.18 & \textbf{0.18} & 0.18 & 0.20 & 0.22 & -19.40 & -23.95 & -23.90 & \textbf{-24.31} & -19.45\\\hline
\multicolumn{1}{|l|}{} & 5x5 & 0.26 & 0.15 & 0.15 & 0.17 & \textbf{0.14} & 0.16 & -9.76 & -17.15 & -15.88 & \textbf{-21.63} & -17.88\\
\multicolumn{1}{|l|}{} & 7x7 & 0.27 & 0.15 & \textbf{0.14} & 0.17 & 0.14 & 0.16 & -11.19 & -21.56 & -17.54 & \textbf{-25.16} & -18.50\\
\multicolumn{1}{|l|}{} & 9x9 & 0.28 & 0.15 & \textbf{0.14} & 0.17 & 0.15 & 0.16 & -12.81 & -24.99 & -19.35 & \textbf{-26.26} & -19.55\\
\multicolumn{1}{|l|}{\multirow{-4}{*}{ZSAD}} & 11x11 & 0.28 & 0.16 & \textbf{0.15} & 0.17 & 0.16 & 0.16 & -14.67 & \textbf{-26.24} & -21.30 & -24.93 & -20.99\\\hline

\multicolumn{1}{|l|}{} & 5x5 & 0.30 & 0.20 & \textbf{0.20} & 0.20 & 0.27 & 0.37 & -15.84 & -18.72 & -19.57 & -14.75 & \textbf{-21.52} \\
\multicolumn{1}{|l|}{} & 7x7 & 0.30 & 0.19 & \textbf{0.19} & 0.20 & 0.28 & 0.38 & -17.05 & -20.70 & -21.33 & -15.59 & \textbf{-21.34} \\
\multicolumn{1}{|l|}{} & 9x9 & 0.30 & 0.19 & \textbf{0.19} & 0.20 & 0.28 & 0.39 & -18.20 & -22.61 & \textbf{-23.13} & -15.56 & -20.51 \\
\multicolumn{1}{|l|}{\multirow{-4}{*}{MedS}} & 11x11 & 0.30 & 0.18 & \textbf{0.18} & 0.19 & 0.28 & 0.38 & -19.08 & -22.04 & \textbf{-24.20} & -14.20 & -19.71 \\ \hline

\multicolumn{1}{|l|}{} & 5x5 & 0.25 & 0.16 & 0.16 & \textbf{0.16} & 0.22 & 0.32 & -10.67 & -17.50 & \textbf{-19.57} & -9.16 & -18.43 \\

\multicolumn{1}{|l|}{} & 7x7 & 0.27 & 0.16 & 0.15 & \textbf{0.15} & 0.24 & 0.35 & -12.51 & \textbf{-22.89} & -22.50 & -9.98 & -17.24 \\

\multicolumn{1}{|l|}{} & 9x9 & 0.30 & 0.19 & \textbf{0.19} & 0.20 & 0.28 & 0.39 & -14.44 & \textbf{-25.62} & -23.96 & -10.36 & -16.75 \\

\multicolumn{1}{|l|}{\multirow{-4}{*}{ZMedS}} & 11x11 & 0.28 & 0.16 & 0.15 & \textbf{0.15} & 0.25 & 0.36 & -17.13 & -23.92 & \textbf{-25.33} & -11.03 & -16.62 \\ \hline
\end{tabular}
\end{adjustbox}
\caption{Aggregate results of studied formulae for the Middlebury stereo dataset (1D patch-based matching) for different correlation windows and cost functions. The best performing formula per row for a given metric is bold and the best performing formula overall is in green.}
\label{table:results1d}
\vspace{-5pt}
\end{table*}

We first evaluate the proposed formulae for one dimensional subpixel refinement. To this end, we use the 2014 Middlebury stereo dataset ~\cite{scharstein2014high}, which contains $10$ high-resolution real image pairs with their corresponding very high-quality ground truth. This high quality ground truth is why we choose this dataset for evaluating subpixel refinement methods on real data. In 2021 the authors of the Middlebury 2014 datasets added 24 images pairs from 11 scenes using a subset of the pipeline of the 2014 dataset \cite{middlebury2021}. We also used the 2021 dataset for evaluation but, as it is still not as widespread in the community as the 2014 dataset, we kept the results apart and discuss them in the Supplementary Material. We have also used the frames from the Active-Passive SimStereo dataset \cite{NEURIPS2022_bc3a68a2}, a simulated dataset, to provide a comparative analysis in a distinct domain. The findings from this additional dataset are thoroughly discussed in Section IV of the Supplementary Material. This inclusion helps to enrich our analysis by offering insights into the performance of our methods across different types of data.

The Middlebury dataset provides image pairs that either feature or lack lighting differences between the left and right images, and it includes images with either imperfect or perfect rectification (where perfect rectification eliminates vertical disparity). In our study, we selected the version with consistent lighting conditions and perfect rectification. This choice was made to prevent undermining cost functions that are sensitive to variations in lighting conditions between the views (\ie all except ZNCC) We evaluate the proposed formulae for each matching function ((Z)NCC, (Z)SSD, (Z)SAD). For comparison, we also evaluate the approximation algorithm for the MedS cost function \cite{LMedSqFlow}. We evaluate when interpolating separately to the left and to the right  (using the barycentric interpolation approach presented in Section~\ref{sec:imageInterpFormulae}), and then when considering both the left and right feature vectors at once (using predictive interpolation as proposed in Section~\ref{sec:formulae_higher_dims:feature_interp}). We compare against the traditional equiangular (Eq.~\eqref{eq:equiangular_refinement}) and parabola (Eq.~\eqref{eq:parabola_refinement}) formulae, as well as against the formula proposed by Shimizu and Okutomi~\cite{Shimizu2005} to cancel the pixel locking effect of parabola-based refinement. We performed the experiment for four different  correlation window sizes, namely  $5\times5$, $7\times7$, $9\times9$ and $11\times11$. To increase the accuracy of the ground truth, we used the Quarter (Q) resolution of the Middlebury dataset.

We first compute the raw disparity map $\nround{\disparity}$ using the same cost or score function as the one used for feature space refinement. We then evaluate the proposed formulae on the pixels that have been successfully matched by the previous step, hereinafter referred to as inliers. As a result, we can measure the actual subpixel accuracy of the refinement formulae rather than the initial matching algorithm's performance, since mismatched pixels would otherwise dominate the error metrics. We report in Table~\ref{table:results1d} the Mean Absolute Error (MAE) between the ground truth disparity $\coord{\check{d}}$ and the refined disparity $\coord{\hat{d}}$, averaged across all images. It is defined as:
\begin{equation}
    MAE = \dfrac{1}{\lvert P_{matched} \rvert} \sum_{\pixel \in P_{matched}} \lvert \coord{\hat{d}}_{[\coord p ]} - \coord{\check{d}}_{[\coord p ]} \rvert.
\end{equation}

\noi In addition to these results, which are aggregated over the entire data set, we provide in the Supplementary Material the MAE and the Root Mean Square Error (RMSE) computed on each individual image.

We also evaluated how much each method is impacted by the pixel locking effect. To this end, we use the Signal-to-Noise Ratio (SNR) where the signal under consideration  $\epsilon(i)$ is the expected subpixel alignment error knowing the disparity fractional part minus the expected subpixel alignment error. It is defined as:
\begin{equation}
    \epsilon(i) = \expec{\hat{d}_i \mid \nround{d_i} - \check{d}_i}{\hat{d}_i - \check{d}_i} - \expec{\hat{d}}{\hat{d} - \check{d}} .
\end{equation}

\noindent The SNR can then be computed as:
\begin{equation}
    SNR = \dfrac{ \sum_{i \in M} \epsilon(i)^2  }{ \sum_{i \in M} (\hat{d}_i - \check{d}_i - \epsilon(i))^2 } .
\end{equation}

\noindent In practice, $\expec{\hat{d}_i \mid \nround{d} - \check{d}_i}{\hat{d}_i - \check{d}_i}$ is approximated by computing the mean subpixel error in discrete intervals of disparity fractional parts; see Fig.~\ref{fig:fract_part_correllation}. We use bins of size $1/40$ pixels, which is a good compromise between precision and statistical significance.

Note that unlike signal reconstruction applications, where a higher SNR means better performance, here the SNR should be as low as possible, meaning the subpixel alignment error cannot be predicted by knowing the fractional part of the disparity.

\subsubsection{Accuracy of subpixel refinement}

\begin{figure*}[t]
    \centering
    
    \begin{subfigure}[t]{0.3\textwidth}
        \includegraphics[width=\textwidth]{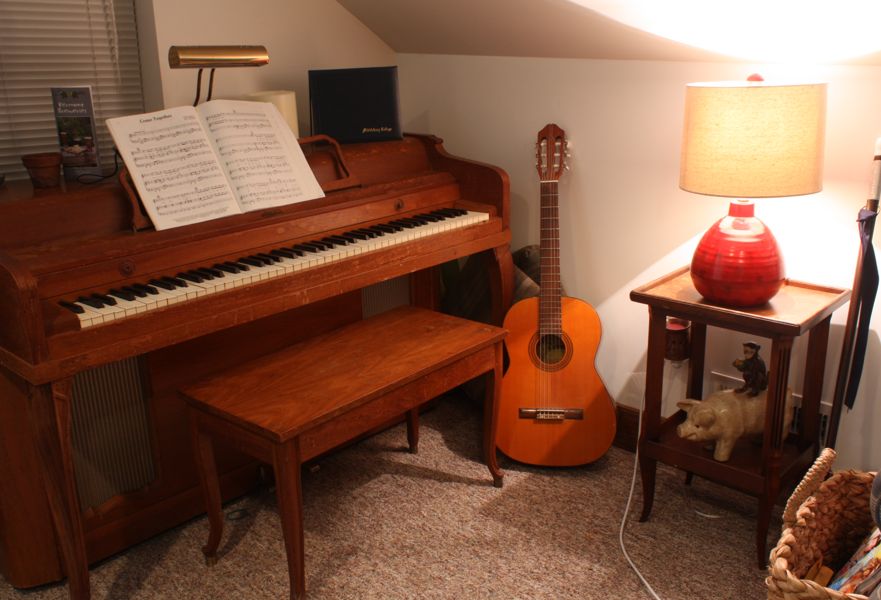}
        \caption{Original image from the Middlebury dataset.}
        \label{fig:pixel_locking_reduction:img}
    \end{subfigure}
    \hfill
    \begin{subfigure}[t]{0.3\textwidth}
        \includegraphics[width=\textwidth]{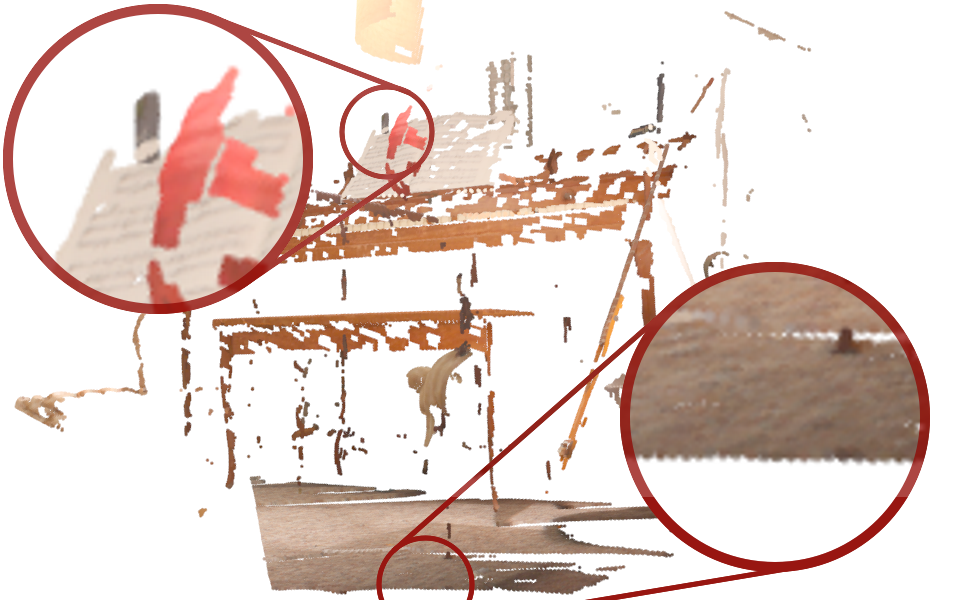}
        \caption{Ground truth point cloud for inliers (sideview).}
        \label{fig:pixel_locking_reduction:gt}
    \end{subfigure}
    \hfill
    \begin{subfigure}[t]{0.3\textwidth}
        \includegraphics[width=\textwidth]{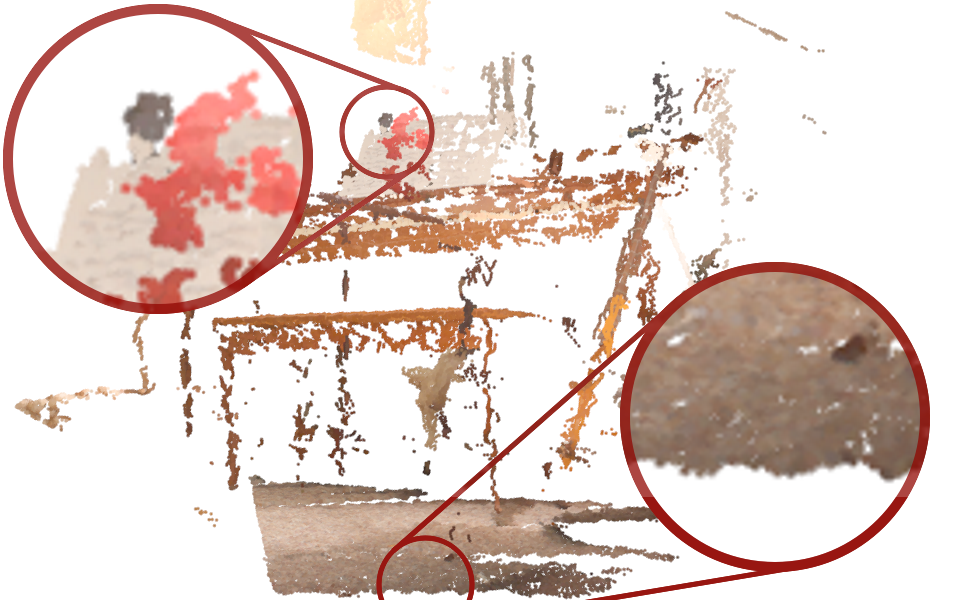}
        \caption{Parabola cost based refined point cloud (sideview).}
        \label{fig:pixel_locking_reduction:cost_based}
    \end{subfigure}
    
    \vspace{5pt}
    
    \begin{subfigure}[t]{0.3\textwidth}
        \includegraphics[width=\textwidth]{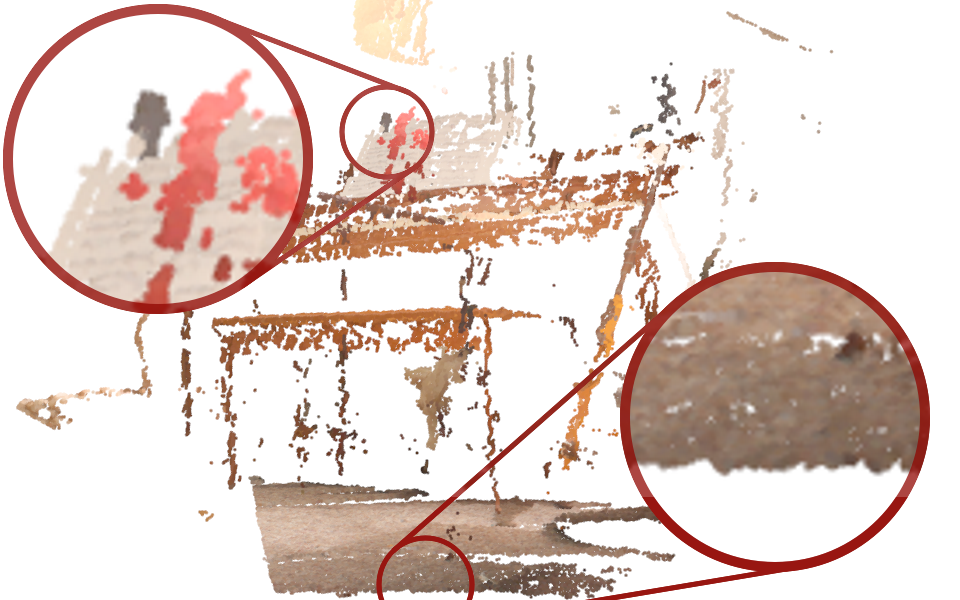}
        \caption{Feature based refined point cloud (sideview).}
        \label{fig:pixel_locking_reduction:feature_based}
    \end{subfigure}
    \hfill
    \begin{subfigure}[t]{0.3\textwidth}
        \includegraphics[width=\textwidth]{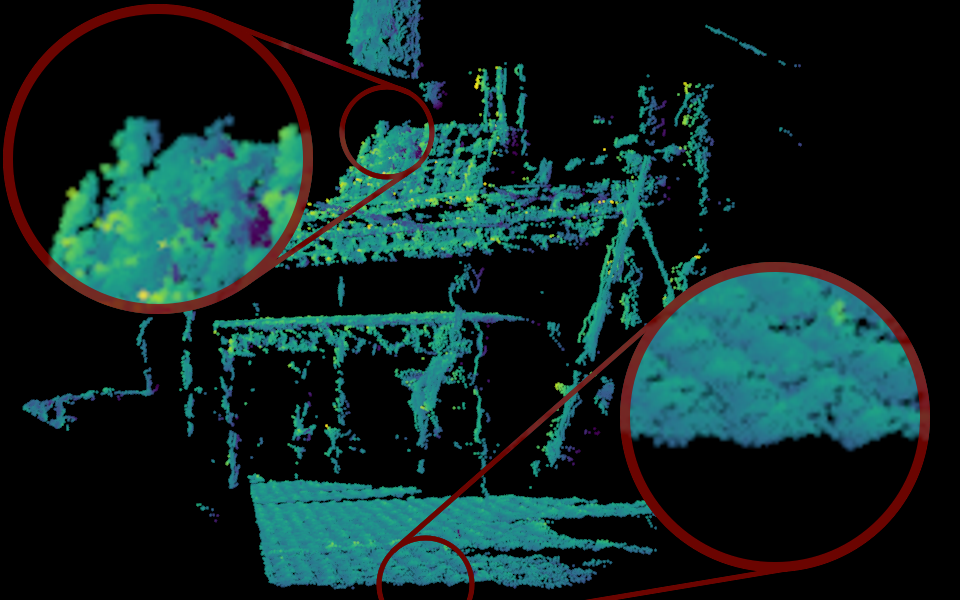}
        \caption{Parabola cost based refined disparity error (sideview).}
        \label{fig:pixel_locking_reduction:cost_based_errors}
    \end{subfigure}
    \hfill
    \begin{subfigure}[t]{0.3\textwidth}
        \includegraphics[width=\textwidth]{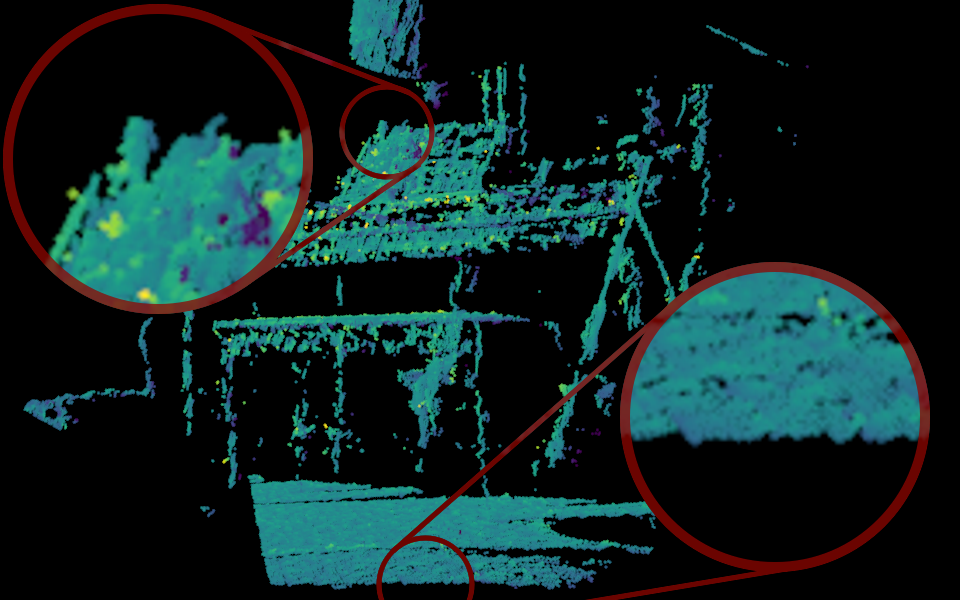}
        \caption{Feature based refined disparity error (sideview).}
        \label{fig:pixel_locking_reduction:feature_based_errors}
    \end{subfigure}
    
    \caption{The benefits of reducing the pixel locking effect: when refining with the parabola based formula, waves can be seen when looking at the 3D point cloud reconstructed from the disparity map when seen from the side (\subref{fig:pixel_locking_reduction:cost_based}). The effect is far less pronounced when using feature based refinement (\subref{fig:pixel_locking_reduction:feature_based}). Here showcased on the piano image from the Middlebury 2014 stereo dataset.}
    \label{fig:pixel_locking_reduction}
\end{figure*}

\begin{figure*}[t]
    \centering
    
    \begin{subfigure}[t]{0.23\textwidth}
        \includegraphics[width=0.85\textwidth]{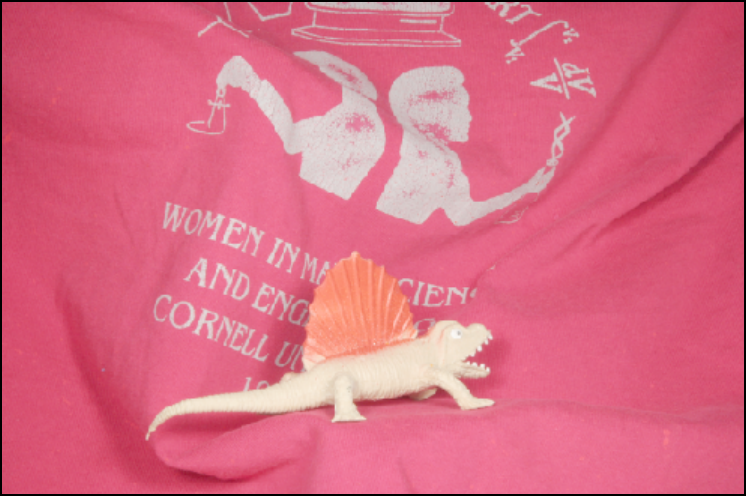}
        \caption{Source image}
        \label{fig:ofview:imgsource}
    \end{subfigure}
    \begin{subfigure}[t]{0.23\textwidth}
        \includegraphics[width=0.85\textwidth]{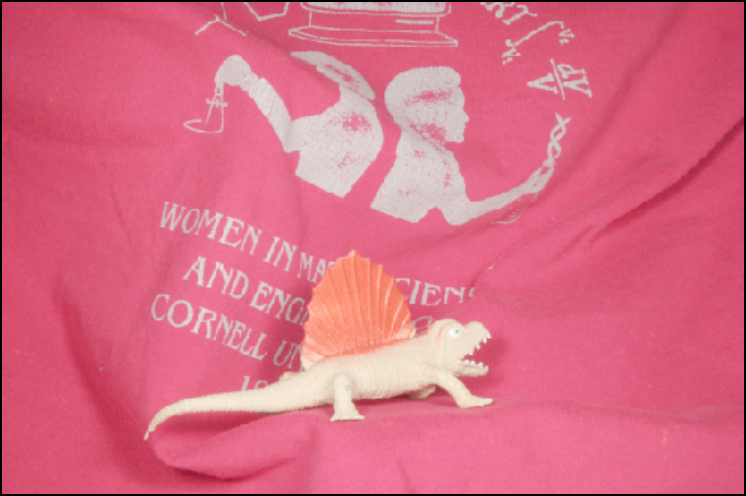}
        \caption{Target image}
        \label{fig:ofview:imgtarget}
    \end{subfigure}
    \hspace{0.02\textwidth}
    \begin{subfigure}[t]{0.23\textwidth}
        \includegraphics[width=\textwidth]{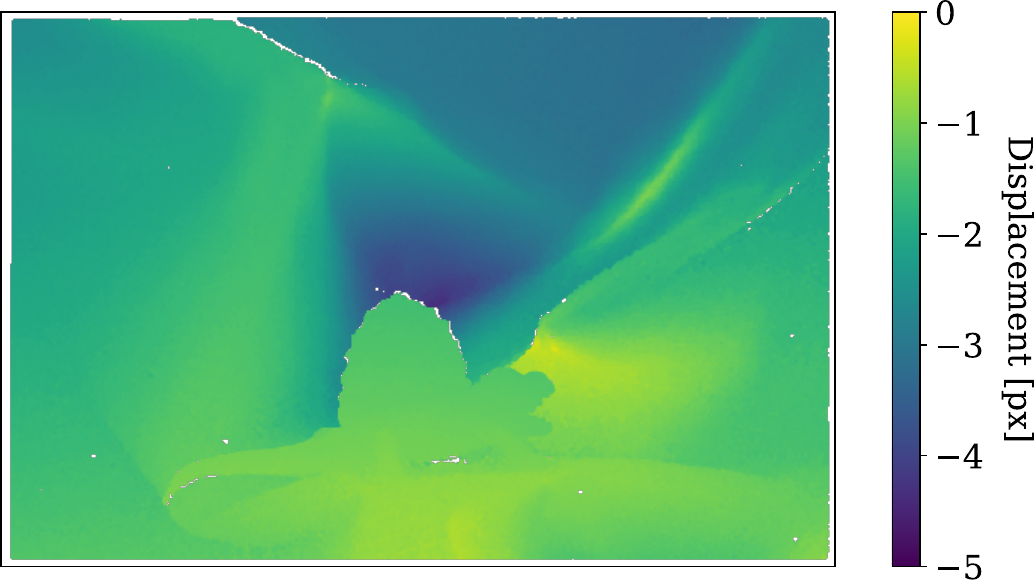}
        \caption{Horizontal Ground Truth}
        \label{fig:ofview:gt_hrz}
    \end{subfigure}
    \begin{subfigure}[t]{0.23\textwidth}
        \includegraphics[width=\textwidth]{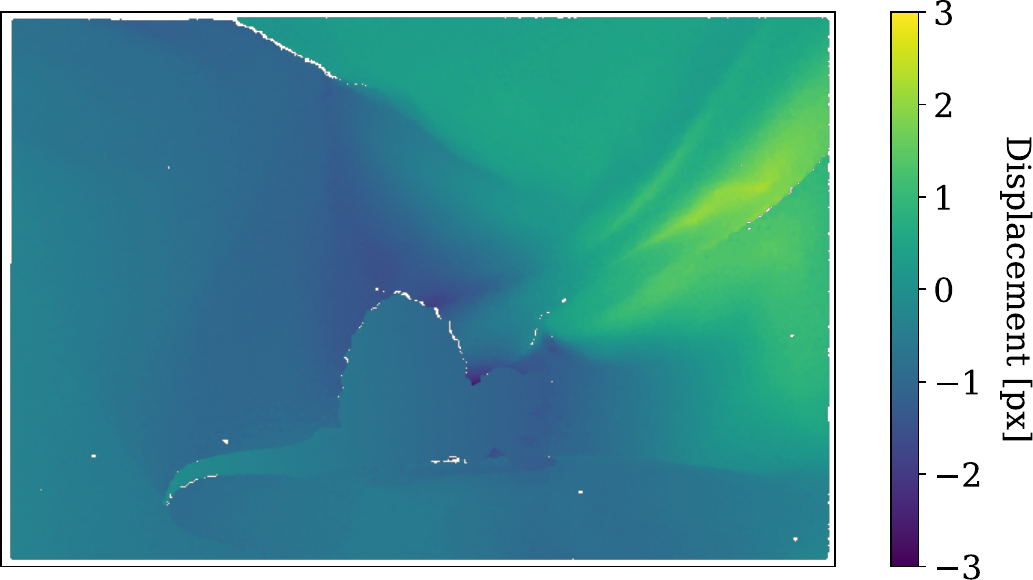}
        \caption{Vertical Ground Truth}
        \label{fig:ofview:gt_vrt}
    \end{subfigure}
    
    \vspace{7pt}
    
    \begin{subfigure}[t]{0.23\textwidth}
        \includegraphics[width=\textwidth]{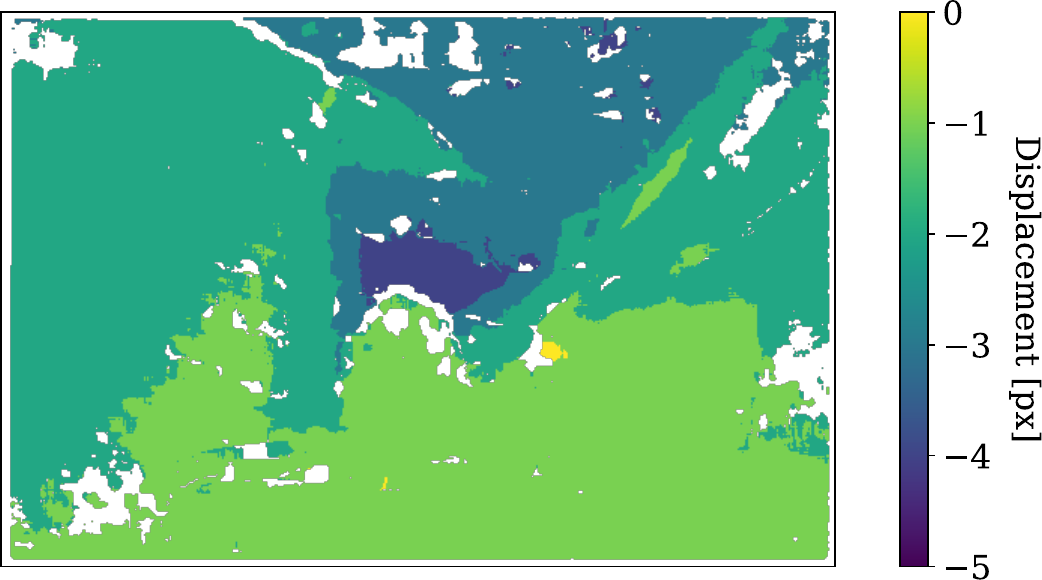}
        \caption{Horizontal raw}
        \label{fig:ofview:raw_hrz}
    \end{subfigure}
    \begin{subfigure}[t]{0.23\textwidth}
        \includegraphics[width=\textwidth]{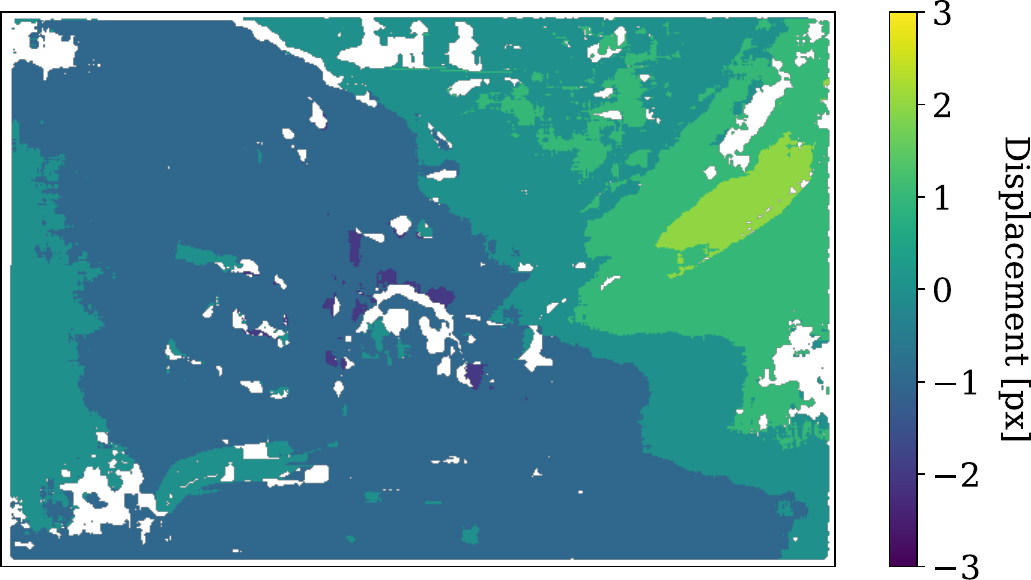}
        \caption{Vertical raw}
        \label{fig:ofview:raw_vrt}
    \end{subfigure}
    \hspace{0.02\textwidth}
    \begin{subfigure}[t]{0.23\textwidth}
        \includegraphics[width=\textwidth]{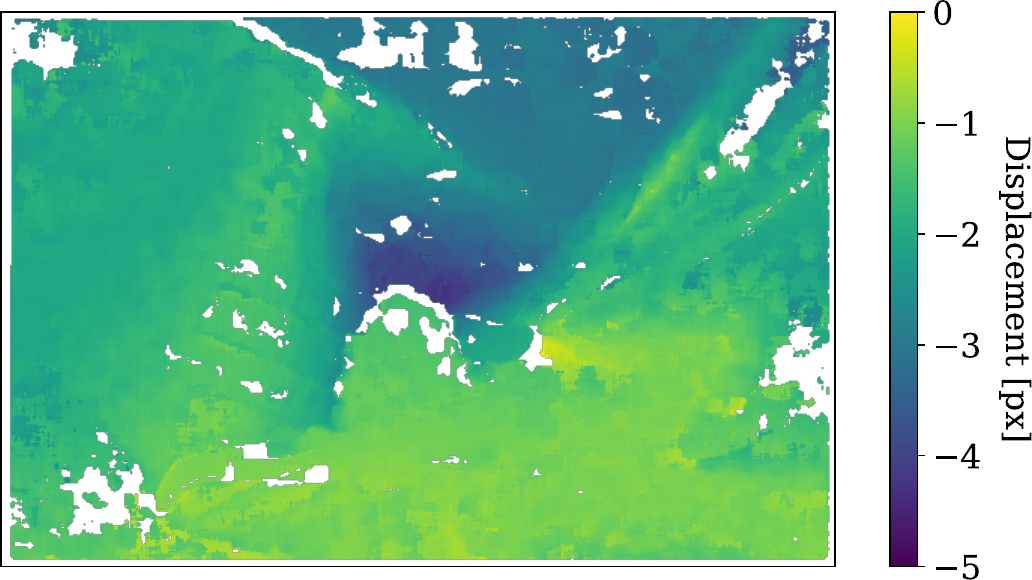}
        \caption{Horizontal Queen refined}
        \label{fig:ofview:queen_hrz}
    \end{subfigure}
    \begin{subfigure}[t]{0.23\textwidth}
        \includegraphics[width=\textwidth]{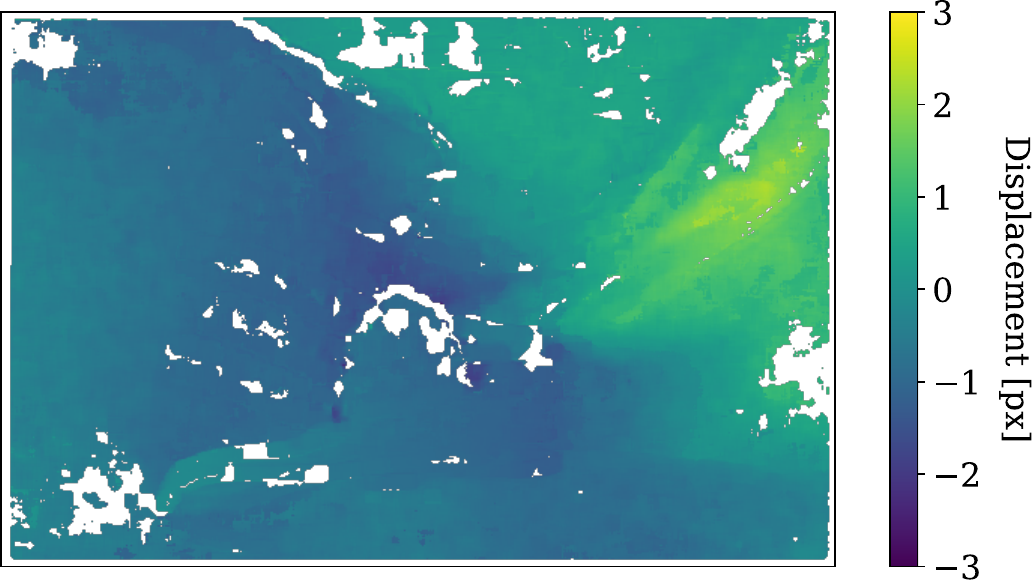}
        \caption{Vertical Queen refined}
        \label{fig:ofview:queen_vrt}
    \end{subfigure}
    
    \caption{Visual results of the Queen contiguity feature based refinement formula on the dimetrodon image from the Middlebury optical flow dataset.}
    \label{fig:ofview}
\end{figure*}

\begin{table*}[t]
\centering
\begin{adjustbox}{max width=\textwidth}
\begin{tabular}{ll|rrrrrrrrrr|}
\cline{3-12}
 &  & \multicolumn{10}{c|}{Mean distance over Middlebury optical flow dataset {[}px{]}} \\ \cline{3-12} 
 &  & \multicolumn{1}{l|}{} & \multicolumn{2}{c|}{Isotropic} & \multicolumn{2}{c|}{Anisotropic} & \multicolumn{1}{l|}{} & \multicolumn{2}{c|}{Features split} & \multicolumn{2}{c|}{Features symmetric} \\ \hline
\multicolumn{1}{|l|}{$C$} & $W$ & \multicolumn{1}{c|}{Raw} & \multicolumn{1}{c|}{Parabola} & \multicolumn{1}{c|}{Equiangular} & \multicolumn{1}{c|}{Parabola} & \multicolumn{1}{c|}{Equiangular} & \multicolumn{1}{c|}{Paraboloid} & \multicolumn{1}{c|}{Rook} & \multicolumn{1}{c|}{Queen} & \multicolumn{1}{c|}{Rook} & \multicolumn{1}{c|}{Queen} \\ \hline

\multicolumn{1}{|l|}{} & 5x5 & 0.36 & 0.26 & 0.28 & 0.34 & 0.35 & 0.34 & 0.20 & \textbf{0.20} & 0.22 & 0.28 \\
\multicolumn{1}{|l|}{} & 7x7 & 0.36 & 0.24 & 0.25 & 0.30 & 0.30 & 0.29 & 0.18 & \textbf{0.17} & 0.20 & 0.24 \\
\multicolumn{1}{|l|}{} & 9x9 & 0.37 & 0.23 & 0.24 & 0.28 & 0.28 & 0.26 & 0.18 & \textbf{0.16} & 0.20 & 0.28 \\
\multicolumn{1}{|l|}{\multirow{-4}{*}{NCC}} & 11x11 & 0.37 & 0.22 & 0.24 & 0.27 & 0.27 & 0.24 & 0.18 & \textbf{0.16} & 0.20 & 0.22 \\ \hline
\multicolumn{1}{|l|}{} & 5x5 & 0.36 & 0.26 & 0.28 & 0.34 & 0.34 & 0.33 & \textbf{0.19} & 0.19 & 0.23 & 0.29 \\
\multicolumn{1}{|l|}{} & 7x7 & 0.36 & 0.27 & 0.25 & 0.30 & 0.30 & 0.28 & 0.18 & \textbf{0.17} & 0.21 & 0.25 \\
\multicolumn{1}{|l|}{} & 9x9 & 0.37 & 0.23 & 0.24 & 0.28 & 0.28 & 0.25 & 0.18 & \textbf{0.16} & 0.20 & 0.23 \\
\multicolumn{1}{|l|}{\multirow{-4}{*}{ZNCC}} & 11x11 & 0.37 & 0.22 & 0.23 & 0.27 & 0.27 & 0.24 & 0.17 & {\color[HTML]{006600} \textbf{0.16}} & 0.20 & 0.22 \\ \hline
\multicolumn{1}{|l|}{} & 5x5 & 0.37 & 0.26 & 0.28 & 0.34 & 0.35 & 0.33 & 0.20 & \textbf{0.20} & 0.23 & 0.28 \\
\multicolumn{1}{|l|}{} & 7x7 & 0.37 & 0.24 & 0.26 & 0.30 & 0.31 & 0.29 & 0.19 & \textbf{0.18} & 0.21 & 0.24 \\
\multicolumn{1}{|l|}{} & 9x9 & 0.37 & 0.23 & 0.24 & 0.28 & 0.28 & 0.26 & 0.18 & \textbf{0.17} & 0.20 & 0.23 \\
\multicolumn{1}{|l|}{\multirow{-4}{*}{SSD}} & 11x11 & 0.37 & 0.23 & 0.24 & 0.27 & 0.27 & 0.24 & 0.18 & \textbf{0.16} & 0.20 & 0.22 \\ \hline
\multicolumn{1}{|l|}{} & 5x5 & 0.36 & 0.26 & 0.28 & 0.34 & 0.34 & 0.34 & 0.24 & \textbf{0.23} & 0.26 & 0.32 \\
\multicolumn{1}{|l|}{} & 7x7 & 0.36 & 0.24 & 0.25 & 0.30 & 0.30 & 0.29 & 0.21 & \textbf{0.19} & 0.23 & 0.27 \\
\multicolumn{1}{|l|}{} & 9x9 & 0.37 & 0.23 & 0.24 & 0.28 & 0.28 & 0.26 & 0.19 & \textbf{0.18} & 0.21 & 0.24 \\
\multicolumn{1}{|l|}{\multirow{-4}{*}{ZSSD}} & 11x11 & 0.37 & 0.22 & 0.23 & 0.27 & 0.27 & 0.24 & 0.19 & \textbf{0.17} & 0.21 & 0.23 \\ \hline
\multicolumn{1}{|l|}{} & 5x5 & 0.37 & 0.27 & 0.27 & 0.35 & 0.34 & 0.30 & 0.21 & \textbf{0.21} & 0.24 & 0.29 \\
\multicolumn{1}{|l|}{} & 7x7 & 0.37 & 0.25 & 0.25 & 0.30 & 0.30 & 0.26 & 0.19 & \textbf{0.18} & 0.21 & 0.25 \\
\multicolumn{1}{|l|}{} & 9x9 & 0.37 & 0.24 & 0.23 & 0.28 & 0.28 & 0.24 & 0.18 & \textbf{0.17} & 0.20 & 0.23 \\
\multicolumn{1}{|l|}{\multirow{-4}{*}{SAD}} & 11x11 & 0.37 & 0.23 & 0.22 & 0.26 & 0.26 & 0.22 & 0.18 & \textbf{0.16} & 0.19 & 0.22 \\ \hline
\multicolumn{1}{|l|}{} & 5x5 & 0.36 & 0.26 & 0.26 & 0.35 & 0.35 & 0.31 & 0.25 & \textbf{0.24} & 0.28 & 0.33 \\
\multicolumn{1}{|l|}{} & 7x7 & 0.36 & 0.25 & 0.24 & 0.30 & 0.30 & 0.26 & 0.21 & \textbf{0.20} & 0.23 & 0.27 \\
\multicolumn{1}{|l|}{} & 9x9 & 0.36 & 0.24 & 0.23 & 0.28 & 0.28 & 0.24 & 0.20 & \textbf{0.18} & 0.21 & 0.24 \\
\multicolumn{1}{|l|}{\multirow{-4}{*}{ZSAD}} & 11x11 & 0.37 & 0.23 & 0.22 & 0.27 & 0.26 & 0.22 & 0.19 & \textbf{0.17} & 0.21 & 0.23 \\ \hline
\multicolumn{1}{|l|}{} & 5x5 & 0.37 & \textbf{0.30} & 0.33 & 0.40 & 0.41 & 0.37 & 0.34 & 0.35 & 0.36 & 0.37 \\
\multicolumn{1}{|l|}{} & 7x7 & 0.37 & \textbf{0.27} & 0.30 & 0.36 & 0.36 & 0.32 & 0.34 & 0.35 & 0.36 & 0.37 \\
\multicolumn{1}{|l|}{} & 9x9 & 0.37 & \textbf{0.26} & 0.29 & 0.33 & 0.34 & 0.29 & 0.33 & 0.35 & 0.36 & 0.36 \\
\multicolumn{1}{|l|}{\multirow{-4}{*}{MedS}} & 11x11 & 0.37 & \textbf{0.25} & 0.27 & 0.31 & 0.32 & 0.27 & 0.33 & 0.34 & 0.36 & 0.36 \\ \hline
\multicolumn{1}{|l|}{} & 5x5 & 0.36 & \textbf{0.29} & 0.31 & 0.40 & 0.40 & 0.36 & 0.33 & 0.34 & 0.36 & 0.36 \\
\multicolumn{1}{|l|}{} & 7x7 & 0.37 & \textbf{0.27} & 0.28 & 0.35 & 0.35 & 0.32 & 0.33 & 0.34 & 0.36 & 0.37 \\
\multicolumn{1}{|l|}{} & 9x9 & 0.37 & \textbf{0.26} & 0.26 & 0.32 & 0.32 & 0.29 & 0.33 & 0.34 & 0.36 & 0.37 \\
\multicolumn{1}{|l|}{\multirow{-4}{*}{ZMedS}} & 11x11 & 0.37 & 0.25 & \textbf{0.25} & 0.30 & 0.30 & 0.27 & 0.33 & 0.34 & 0.36 & 0.37 \\ \hline
\end{tabular}
\end{adjustbox}
\caption{Aggregate results of the studied formulae for the Middlebury optical flow dataset (2D patch-based matching) for different correlation windows and cost functions. The best performing formula per row is in bold and the overall best performing formula is in green.}
\label{table:results2d}
\vspace{-5pt}
\end{table*}

The results of our experiment, reported in Table~\ref{table:results1d}, show a clear benefit of the feature space interpolation formula when using Barycentric interpolation, as the peak performance obtained with this approach (a MAE of $0.124$px) is way better than the peak performance of any other approach. Especially the additional uncertainty brought by the larger number of feature vectors considered is too detrimental when using predictive interpolation.

One important point to notice is that the proposed formulae seem to be much more impacted by the size of the correlation windows than the cost-volume based formulae. When using the NCC cost function and a large correlation window ($9\times9$ and above), the method of Shimizu and Okutomi \cite{Shimizu2005} has superior performances. The results in Table~\ref{table:results1d} indicate that other cost functions follow a similar trend.  This is probably due to the fact that the disparities within the correlation window are not uniform, creating misalignment within the patches when they are matched \cite{bleyer2011patchmatch}. This is, however, not an issue for cost function-based interpolation methods since the generated noise will be symmetric on each side of the optimum subpixel position, just like the parabola or equiangular kernel. For feature-based refinement, this will pose a problem, which will become more and more visible as the size of the correlation window, and thus the discrepancies between pixels disparities in the window, increases.

Note that feature-based interpolation is not suitable for the SAD cost function, with the equiangular cost-based refinement formula  outperforming  all alternative formulae for this cost function. However, the performance of any combination of method and correlation windows on SAD is still far from the best performing combination.

The results of Kim and Kak's \cite{LMedSqFlow} image interpolation for MedS and ZMedS costs are very poor, but remember that their iterative algorithm \cite{LMedSqFlow} with which we are comparing our formulae is an approximation. These two cost functions also performed relatively poorly for the cost interpolation-based methods compared to other cost functions. This is probably because the pixel with the median square error is likely to be a different pixel for different neighboring image patches, causing additional noise even when interpolating in the cost space.

\textbf{The absolute best formula is a feature-based interpolation using a $5\times5$ correlation window and ZNCC cost function.} As a matter of fact, the feature-based interpolation with the ZNCC cost function is the most effective for all correlation window sizes except $11\times11$, for which the SAD with equiangular cost refinement provides the best result. In general, using a smaller correlation window is often not desirable for discrete matching. However, one can use different window sizes for the matching and refinement, since the refinement process is a local process and as such it should not be adversely impacted by ambiguous image patches elsewhere in the image.

\subsubsection{Robustness to the pixel-locking effect}

In terms of sensitivity to the pixel-locking effect, the parabola-based interpolation formula  shows an SNR up to one order of magnitude worse than alternative methods with clearly visible artifacts in the point clouds and depth maps reconstructed from the disparity; see Figure~\ref{fig:pixel_locking_reduction}. The three remaining formulae all have decent performances that are quite similar to one another. This shows that, in almost all cases, parabola cost interpolation should be avoided. The equiangular cost-based interpolation formula or Shimizu and Okutomi \cite{Shimizu2005} cancellation method can always replace their parabola counterpart with minimal performance loss and no overhead in terms of computation time. Feature-based interpolation, while slightly more expensive in terms of computation time, offers good benefits in terms of reconstruction accuracy as well as robustness to the pixel-locking effect.

\subsection{Two dimensional patch-based matching}
\label{sec:experimental:twod}

To test the proposed formulae for 2D subpixel refinement, we use the optical flow Middlebury dataset, a collection of three real and seven simulated image pairs with very high quality optical flow ground truth \cite{baker2011database}. We evaluate, for each matching function ((Z)NCC, (Z)SSD, (Z)SAD), the proposed formulae using both the Rook (Fig.~\ref{fig:interp_2d:cut_rook}) and Queen (Fig.~\ref{fig:interp_2d:cut_queen}) contiguity. We first consider the case where each four corners of $\nround{\vec{d}}$ are considered individually, using barycentric interpolation for the Rook contiguity (Fig.~\ref{fig:interp_2d:cut_rook}) and predictive interpolation for the Queen contiguity (Fig.~\ref{fig:interp_2d:cut_queen}). Then, we consider the case where  each four corners of $\nround{\vec{d}}$ are considered all at the same time, using predictive interpolation in all cases  (Fig.~\ref{fig:interp_2d:sym_rook} and \ref{fig:interp_2d:sym_queen}). We compare against the traditional separable formula (Eq.~\ref{eq:multidimseparable}) for parabola an equiangular cost interpolation, as well as against the anisotropic formula of Shimizu and Okutomi~\cite{2DSimultaneousSupixel}, again both with parabola and equiangular cost and paraboloid fitting (using least squares).

Similar to Section~\ref{sec:experimental:oned}, we first compute the raw disparity map using the same cost function as the one used for subpixel refinement. We then evaluate our formula on the inliers. To be consistent with the $MAE$, which we used in the 1D case (Section~\ref{sec:experimental:oned}), we report the mean norm of the difference between the ground truth optical flow $\coord{\check{d}}$ and the refined optical flow $\coord{\hat{d}}$, denoted as $MD$:
\begin{equation}
    MD = \dfrac{1}{\lvert P_{matched} \rvert} \sum_{\pixel \in P_{matched}} \left\lVert \coord{\hat{d}}_{[\coord p ]} - \coord{\check{d}}_{[\coord p ]} \right\rVert_2.
\end{equation}

\noindent We also provide, in the Supplementary Material, the $MD$ per image.

The results, reported in Table~\ref{table:results2d}, show very clear benefits for feature-based formulae, and more specifically the Queen contiguity-based formula when each corner of $\nround{\vec{d}}$ is considered independently. \textbf{In term of accuracy, the overall best performing method is once again the ZNCC based formula, but this time when using an $11\times11$ correlation window.} In fact, unlike  the one dimensional case, the performance of almost all feature-based formulae increases with the size of the correlation window. This can be explained by multiple factors but the most important one is that when reconstructing optical flow, as long as the relative motion between two consecutive frames is small, the search space is smaller and the risk of having large flow variations within the correlation windows is thus limited.

Visually, the improvement of the proposed feature-based refinement formula is very clear. In Figure~\ref{fig:ofview}, for example, one can see that most of the details invisible in the raw displacement map can be spotted on the refined map.

\section{Conclusion and future work}
\label{sec:conclusion}

This paper studied closed-form formulae for subpixel refinement. We proposed new and extended existing feature-based refinement formulae for any-dimensional patch-based matching. We have evaluated those formulae against existing closed-form formulae based on cost interpolation for stereo matching and optical flow evaluation. Our experiments show that the proposed formulae offer clear benefits in almost all of the considered cases. In particular, our generalized two dimensional formula for the ZNCC cost function attains the absolute best performance compared to all the other tested formulae on the Middlebury optical-flow dataset. The proposed feature-based formulae can easily and directly replace their cost-based counterparts in existing machine vision and image processing applications. While the focus of this work was mostly mathematical, \ie deriving and analyzing formulae, we plan in the future to consider practical applications by implementing the proposed formulae in more complete pipelines for applications such as  stereo vision. We also plan to generalize the proposed formulae to the non-linear features used in practice, for which interpolation in the image space is no longer equivalent to interpolation in the feature space.

\section{Acknowledgments}
This research is supported by the Australian Research Council  {https://www.arc.gov.au/} (Grant no. DP210101682).

{\small
\bibliographystyle{ieee_fullname}
\bibliography{bibliography}
}

\end{document}


\title{Generalized Closed-form Formulae for Feature-based Subpixel Alignment in Patch-based Matching -- Supplementary material}

\author{Laurent Valentin Jospin\textsuperscript{1,2}, Hamid Laga\textsuperscript{3}, Farid Boussaid\textsuperscript{2}, Mohammed Bennamoun\textsuperscript{2},~\IEEEmembership{Senior Member,~IEEE,}\\
\textsuperscript{1}École polytechnique fédérale de Lausanne,\textsuperscript{2}University of Western Australia, \textsuperscript{3}Murdoch University\\
{\tt\small laurent.jospin@epfl.ch}, {\tt\small  H.Laga@murdoch.edu.au},\\{\tt\small  \{farid.boussaid,mohammed.bennamoun\}@uwa.edu.au}}

\maketitle

\newcommand{\shortauthors}{Jospin et al.}

\section{An argument in favor of interpolation in image space over cost space}
\label{seq:whyimageinterpisbetter}

The image measurement process for digital cameras, to avoid aliasing problem, an optical low pass filter \cite{Hosseini07} to remove the high frequencies in the input image signal before sampling the values for each pixels. The image signals can thus be considered as low pass signals. Moreover, it is reasonable to assume images are bounded signals.

\begin{proposition}
\label{prop:bernsteininequality}
The absolute value of the derivative of a band limited and bounded signal $f(x)$ is bounded by a constant which increase linearily with the cutoff frequency $\omega_c$ and the bound $b$ such that $|f(x)| < b \forall x$.
\end{proposition}

\begin{proof}

The proof we give here will follow a similar procedure to the proof given by Zygmund \cite{zygmund1932remark} for trigonometric polynomials of finite degree and the proof given by Pinsky \cite{pinsky2008introduction} for band limited functions.

First notice that, because of the Nyquist–Shannon sampling theorem and the commutativity of the convolution product we have:

\begin{equation}
\label{eq:NyquistShannonDerivConv}
    \dfrac{\partial f}{\partial x}(x) = \left( \text{\textcyr{\CYRSH}}_{1/a} (x) \cdot f(x) \right) \ast \dfrac{\partial sinc\left( a x\right)}{\partial x} = a \sum_{n=-\infty}^{\infty} f\left(\dfrac{n}{a}\right) \left( \dfrac{\pi \left(ax-n\right) cos\left(\pi (ax-n)\right) - sin \left(\pi (a x-n)\right)}{\pi (ax-n)^2} \right),
\end{equation}

\noindent with $a = \dfrac{\omega_c}{\pi}$, $\text{\textcyr{\CYRSH}}_{1/a} (x) = \sum_{n=-\infty}^{\infty} \delta_{n/a} (x)$ and $sinc(x) = \dfrac{sin(\pi x)}{\pi x}$.\\

From Equation~\ref{eq:NyquistShannonDerivConv}, one can set an upper bound on the value of $\left|\dfrac{\partial f}{\partial x}\left( \dfrac{1}{2a}\right) \right|$:

\begin{equation}
\label{eq:derivative_band_limited_limit}
    \left|\dfrac{\partial f}{\partial x}\left( \dfrac{1}{2a}\right)\right| \leq \dfrac{4 ab}{\pi} \sum_{n=-\infty}^{\infty} \dfrac{1}{(2n - 1)^2} = \dfrac{4 \omega_c b}{\pi^2} \left(\dfrac{6}{4} \cdot \zeta(2)\right) = \omega_c b,
\end{equation}

\noindent with $\zeta(z)$ being the Riemman zeta function. The solution to the Basel problem, found by Leonard Euler, states that $\zeta(2) = \pi^2/6$. This upper bound generalize to all possible $x$ by substituting any $g_k(x) = f(x + k), k \in \Reals$ to $f(x)$, since all $g_k(x)$ have the same bound $b$ as $f(x)$ and their Fourier transform has the same support. The bound $\omega_c b$ is tight, as we have equality for all $x \in \mathbb{R}$ for any function of the form $b e^{i\omega_c x}$. This result is often referred to in the signal processing community as Bernstein's inequality, even if the original result by Bernstein \cite{bernstein1912equations} was about Fourier series and not Fourier transforms.

\end{proof}

\begin{corollary}
\label{prop:bernsteininequalitycorollary}
The total absolute variation of such a band limited signal across a distance $\Delta x$ is bounded by a similar constant:

\begin{equation}
    \label{eq:bernsteininequalitycorollary}
    | f(x) - f(x + \Delta x) | \leq \omega_c b \left| \Delta x\right|.
\end{equation}

\end{corollary}

\begin{proof}
 The proof immediately derive from Equation~\eqref{eq:derivative_band_limited_limit}.
\end{proof}

It follows from Corollary~\ref{prop:bernsteininequalitycorollary} that the range of values a band limited signal can take at a given position between two samples value is bounded, and the size of the gap is proportional to the cutoff frequency. 

This, in turn, motivates the choice of image or feature space interpolation over cost volume interpolation. Nyquist–Shannon interpolation is not practical in applications and the interpolation scheme used has a certain uncertainty associated with it which is correlated to the range of possible values the signal can take. To see why in matters for subpixel refinement, lets evaluate, for example, the SSD cost curve (\ie cost curve obtained from the continuous image signal ). First, the differences between the pixels values need to be computed. They form a collection of signals (as functions of the disparity $\disparity$) designated as $\Delta F (\disparity)$:

\begin{equation}
    \Delta F (\disparity) = \tensor{F}_{s}(\pixel,.) - \tensor{F}_{t}(\pixel + \disparity,.).
\end{equation}

\noindent Since the target images pixels are band limited and bounded, all signals in $\Delta F (\disparity)$ are band limited (with the same cutoff frequency as the target image) and bounded (as $\tensor{F}_{s}(\pixel,i)$ is a finite constant vector, and adding a finite constant to a bounded signal cannot make it unbounded or change its cutoff frequency). In what follows, $b$ designate the upper bound of any component of $\Delta F (\disparity)$, \ie $\forall i \in [1, |c_{1{\tensor{F}}}|] \left| \Delta F_i (\disparity)\right| \leq b \forall \disparity$.

The cost signal is the the sum of squares of the components of $\Delta F (\disparity)$:

\begin{equation}
    SSD(\disparity) = \sum_{i=0}^{|c_{1{\tensor{F}}}|} (\Delta F (\disparity))^2,
\end{equation}

\noindent which is a sum of squares of band-limited signals. The product of two band-limited signals with cutoff frequency $\omega_{c}$ has a cutoff frequency $2\omega_{c}$. 

Now, assuming that the images have been sampled at the Nyquist rate, we have $1px = \dfrac{\pi}{\omega_{c}}$. It means the furthest away a disparity value can be from a sample is equal to half a pixel, or $\dfrac{\pi}{2 \cdot \omega_{c}}$. According to Equation~\eqref{eq:bernsteininequalitycorollary}, the upper bound on the variability of $\Delta F (\disparity)$ is $\dfrac{\pi b}{2}$. Squaring it and multiplying by the number of features $|c_{1{\tensor{F}}}|$ yield an upper bound for the cost signal variation:

\begin{equation}
    | SSD(\disparity) - SSD(\disparity + 0.5px) | \leq |c_{1{\tensor{F}}}| \left(\dfrac{\pi b}{2}\right)^2.
\end{equation}

\noindent This is smaller than the estimate of $|c_{1{\tensor{F}}}| \pi b^2$, obtained by only applying Equation~\eqref{eq:bernsteininequalitycorollary} to a sum of $|c_{1{\tensor{F}}}|$ signals with cutoff frequency $2\omega_{c}$ and bound $b^2$. This shows that interpolating using only the cost values ignores valuable information compared to interpolating in the image domain. In theory, with perfect Nyquist–Shannon interpolation and when using the SSD cost, interpolating the images at intervals of 0.5 pixels is sufficient. The interpolation can then be performed in the cost space, but as stated just above, Nyquist–Shannon interpolation is not practical in applications.

Similar arguments apply for the SAD cost, which takes an absolute value, which is akin to multiplying the signal by its sign function (which does not have a band-limited spectrum and thus makes the cost signal band illimited), as well as the NCC cost, which normalize the signal. In fact, the SAD and NCC cost functions and their variants will yield a cost signal with potentially infinite support in the Fourier domain. 

By extension, it is, a priory, preferable to interpolate beforehand in image space than after computing the cost volume.

\section{Error analysis of the image interpolation models}

Section~\ref{seq:whyimageinterpisbetter} provided an argument on why image interpolation should, on average, yield better performances than cost interpolation for arbitrary band-limited signals. While in theory sinc-interpolation would be preferable for such signals, in practice it is infeasible. In fact, to get closed form formulae that require only algebraic operations and, by extension. no complex iterative algorithms like Newton's method, it is necessary to use linear interpolation (or predictive interpolation when enough data points are available). In this section, we will present a statistical error model for linear interpolation and show how the major source of errors are dealt with with the different cost functions usually used for patch based matching.

First, lets observe that basic (\ie non normalized and non zero-mean) cost functions like SSD or SAD assume, given the source signal $s(\disparity)$ and target signal $t(\disparity)$ the following error model:

\begin{equation}
\label{eq:basedcosterrormodel}
    t(\disparity) = s(\disparity) + \varepsilon,
\end{equation}

\noindent where $\varepsilon$ is a random variable with a symmetric distribution and a mean of $0$. If the error model is biased, then so will the estimator. Zero mean cost functions are a bit more lenient and accommodate an error model with a constant bias $c$:

\begin{equation}
\label{eq:zeromeanerrormodel}
    t(\disparity) = s(\disparity) + c + \varepsilon,
\end{equation}

\noindent normalized error functions accommodate an error model with a scaling factor $a$:

\begin{equation}
\label{eq:normalizederrormodel}
    t(\disparity) = a \cdot s(\disparity) + \varepsilon,
\end{equation}

\noindent and zero mean normalized error functions accommodate both:

\begin{equation}
\label{eq:zeromeannormalizederrormodel}
    t(\disparity) = a \cdot s(\disparity) + c + \varepsilon.
\end{equation}

\noindent When doing discrete matching, $a$ and $c$ are due to a difference in gain, respectively bias, in the image sensors taking the source and target measurements \cite{Scharstein2002}. But when interpolating, additional errors can be caused by the imprecision due to the mismatch between the original source signal $s(\disparity)$ and the approximation $\hat{s}(\disparity)$.

If given a function $f$ and $\hat{f}$ an approximation of $f$ (\eg a linear interpolation between two points of $f$) in an interval which, for convenience, we fix at $1px$ , finding an estimate of the gain $\hat{a}$ and the bias $\hat{c}$ is equivalent to solving the following problem:

\begin{equation}
\label{eq:minzlinearinterperror}
    \left( \begin{array}{c}
        \hat{a} \\
        \hat{c}
    \end{array} \right) = \argmin_{a, c} \dotprod{f(x) - a\hat{f}(x) - c}{f(x) - a\hat{f}(x) - c},
\end{equation}

\noindent where $\dotprod{}{}$ is the functional inner product in the sample interval $[0px,1px)$, defined as:

\begin{equation}
    \dotprod{f(x)}{g(x)} = \int_0^{1} f(x)g(x)dx.
\end{equation}

\noindent Equation~\eqref{eq:minzlinearinterperror} is a quadratic form in $\hat{a}$ and $\hat{c}$, thus solving Equation~\eqref{eq:minzlinearinterperror} is equivalent to solving:

\begin{equation}
\label{eq:minzlinearinterperrorlinearform}
    \matr{H} \left( \begin{array}{c}
        \hat{a} \\
        \hat{c}
    \end{array} \right) = \vec{d},
\end{equation}

\noindent where the Hessian of the system $\matr{H}$ is given by :

\begin{equation}
    \matr{H} = \left( \begin{array}{cc}
        \dotprod{\hat{f}(x)}{\hat{f}(x)} & \dotprod{\hat{f}(x)}{1} \vspace{5pt} \\
        \dotprod{\hat{f}(x)}{1} & \dotprod{1}{1}
    \end{array} \right),
\end{equation}

and the vector $\vec{d}$ is given by:

\begin{equation}
    \vec{d} = \left( \begin{array}{c}
        \dotprod{f(x)}{\hat{f}(x)} \vspace{5pt} \\
        \dotprod{f(x)}{1}
    \end{array} \right).
\end{equation}

solving Equation~\eqref{eq:minzlinearinterperrorlinearform} gives us:

\begin{equation}
\label{eq:estsinsol}
    \left( \begin{array}{c}
        \hat{a} \\
        \hat{c}
    \end{array} \right) =  \dfrac{1}{\dotprod{\hat{f}(x)}{\hat{f}(x)} - \dotprod{\hat{f}(x)}{1}^2} \left( \begin{array}{c}
        \dotprod{f(x)}{\hat{f}(x)} - \dotprod{\hat{f}(x)}{1}\dotprod{f(x)}{1} \vspace{5pt}\\
        \dotprod{\hat{f}(x)}{\hat{f}(x)}\dotprod{f(x)}{1} -\dotprod{f(x)}{\hat{f}(x)}\dotprod{\hat{f}(x)}{1}
        
    \end{array} \right)
\end{equation}

Notice that the contribution $\dotprod{\hat{f}(x)}{\hat{f}(x)} - \dotprod{\hat{f}(x)}{1}^2$ is always positive, due to Jensen's inequality \cite{needham1993visual}. In fact, it is the variance of the linear approximation values. $\dotprod{f(x)}{\hat{f}(x)} - \dotprod{\hat{f}(x)}{1}\dotprod{f(x)}{1}$ on the other hand is the covariance of the function and the approximation. 

\begin{proposition}
\label{prop:sumbiasgainformula}

Given two image signals $f_1$ and $f_2$ from a function Hilbert space $H$ and their two unbiased approximations in a pixel interval: $a_1*\hat{f}_1 + c_1$ and $a_2*\hat{f}_2 + c_2$, with $\hat{f}_1$ and $\hat{f}_2$ being linear interpolations of $f_1$ and $f_2$, then $a_1*\hat{f}_1 + c_1 + a_2*\hat{f}_2 + c_2$ is an unbiased approximation of $f_1 + f_2$.
\end{proposition}

\begin{proof}

Notice that $\hat{f}_1$, $\hat{f}_2$ as well as their corrected versions $a_1*\hat{f}_1 + c_1$ and $a_2*\hat{f}_2 + c_2$ belongs to the Hilbert space $L_{1px}$ defined as:

\begin{equation}
    L_{1px} = \left\{ \lambda_1 (1 - x) + \lambda_2 (x), (\lambda_1, \lambda_2) \in \Reals^2 \right\}.
\end{equation}

\noindent The unbiased approximations $a_1*\hat{f}_1 + c_1$ and $a_2*\hat{f}_2 + c_2$ can be obtained by an orthogonal projection from $H$ to $L_{1px}$. Any orthogonal projection between Hilbert spaces being a linear operator, the optimal approximation of the sum has to be the sum of the optimal linear approximations.

\end{proof}

\begin{proposition}
\label{prop:linearerrorcorrelation_base}

Given an Hilbert space $F$ of bandlimited image signal $f(x)$ with cutoff frequency $\omega_c = \pi px^-1$ and their linear approximation $\hat{f}(x) = ({\textcyr{\CYRSH}}_{1px} (x) f(x)) \ast trig(x)$, where $trig(x)$ is the triangular function, then if an error model similar to the one use in Equation~\eqref{eq:minzlinearinterperror} is considered in each discrete pixel interval, then there exist a function in $F$ such that $ a \neq 1$.

\end{proposition}

\begin{proof}

Consider the family of sinus functions of the form $f(x) = sin(\omega x + \varphi)$ of frequency $\omega < \omega_c$ and arbitrary phase $\varphi \in [0,2\pi)$. The linear approximant is given by $\hat{f}(x) = \dfrac{\omega_c \cdot sin(\varphi)}{\pi}(1-x) + \dfrac{\omega_c \cdot sin(\pi \omega / \omega_c + \varphi)}{\pi} x$.

If we substitute the extract expressions for $f(x)$ and $\hat{f}(x)$ in Equation~\eqref{eq:estsinsol} and simplify, we can get an expression for $\hat{a}$:

\begin{equation}
\label{eq:ahatclosedform}
    \hat{a} = \dfrac{12 \left( \sin \left(\omega + \varphi \right) -\,\sin (\varphi) \right) -6\,\omega\, \left(  \cos \left(
 \omega + \varphi \right) + \cos (\varphi) \right) 
 }{\omega^2\,\left( \sin \left(\omega + \varphi \right)-\,\sin 
 (\varphi) \right)} = \dfrac{12}{\omega^2} - \dfrac{6}{\omega} \dfrac{ \cos \left(
 \omega + \varphi \right) + \cos (\varphi) }{ \sin \left(\omega + \varphi \right)-\,\sin 
 (\varphi) }.
\end{equation}

\noindent The function in Equation~\eqref{eq:ahatclosedform} lead to an indefinite algebraic formulation are the cases where $\varphi = \dfrac{-\omega + \pi}{2} + n\pi$, $n\in \Naturals$. The reason why is that in that case we have $\omega + \dfrac{-\omega + \pi}{2} = -\left(\dfrac{-\omega + \pi}{2}\right) + n\pi$. These correspond to the case where the linear interpolant is flat and the solution becomes ambiguous between $\hat{a}$ and $\hat{c}$. But the limit of the function is well defined and we get:

\begin{equation}
    \lim_{\varphi \to (-\omega + \pi)/2} \hat{a} =  {{6\,\omega \sin \left({{3\,\omega}\over{2}}\right)+12\,\cos \left({{3\,\omega
 }\over{2}}\right)+6\,\omega \sin \left({{\omega}\over{2}}\right)-12\,\cos 
 \left({{\omega}\over{2}}\right)}\over{\omega^2\cos \left({{3\,\omega}\over{2}}
 \right)- \omega^2 \cos \left({{\omega}\over{2}}\right)}}.
\end{equation}

\noindent While this formulation also has a singularity at $\omega = \dfrac{\pi}{2}$, the limit is once again well defined and we have:

\begin{equation}
    \lim_{\omega,\,\varphi \to \pi/2,\,\pi/4} = \dfrac{48}{\pi^2} - \dfrac{12}{\pi}
\end{equation}

An important point is that it means $\hat{a}$ is continuous everywhere. It implies it is constant with respect to the phase $\varphi$, \ie:

\begin{equation}
    \dfrac{\partial \hat{a}}{\partial \varphi} = \dfrac{6\,\sin ^2\left(\omega+\varphi\right)+6\,\cos ^2\left(\omega+\varphi\right)-6\,\sin ^2(\varphi)
 -6\,\cos ^2(\varphi)}{\omega\,\sin ^2\left(\omega+\varphi\right)-2\,\sin (\varphi)\,\omega\,\sin 
 \left(\omega+\varphi\right)+\sin ^2(\varphi)\,\omega} = 0;
\end{equation}

\noindent This allows to simplify the formula for $\hat{a}$ by setting $\varphi = 0$:

\begin{equation}
\label{eq:ahatclosedsimplifiedform}
    \hat{a} = \dfrac{12\,\sin w-6\,w\,\cos w-6\,w}{w^2\,\sin w}
\end{equation}

Notice that the expression given in Equation~\eqref{eq:ahatclosedsimplifiedform} leads to a ratio of the form $0/0$ for $\omega = 0$ and $\omega = \pi$, but the limits are still well defined:

\begin{equation}
\label{eq:ahatclosedsimplifiedformextremum}
    \begin{array}{cc}
        \lim_{\omega \to 0} \hat{a} = 1 ~~~ & ~~~  \lim_{\omega \to \pi} \hat{a} = \dfrac{12}{\pi^2}.
    \end{array}
\end{equation}

Another important fact is that it shows that $\hat{a}$ is always greater than or equal $1$, with equality only when $\omega = 0$. This is because we can choose $\varphi = \pi - \dfrac{\pi}{2} \omega$, which mean of $f(x)$ and $\hat{f}(x)$ are $0$ in the integration interval. By the properties of the sin function, $f(x)$ will always be greater or equal than $\hat{f}(x)$ when $f(x) > 0$ and smaller or equal than $\hat{f}(x)$ when $f(x) \leq 0$. As a consequences we have for $\varphi = \pi - \dfrac{\pi}{2} \omega$:

\begin{equation}
    \hat{a} = \dfrac{\dotprod{f}{\hat{f}}}{\dotprod{\hat{f}}{\hat{f}}}, ~ \dotprod{f}{\hat{f}} > \dotprod{\hat{f}}{\hat{f}}.
\end{equation}

\noindent Which, in turn, demonstrates that:

\begin{equation}
     \hat{a} \geq 1 ~ \forall f(x) \in \left\{sin(\omega x + \varphi) | \omega \in [0,\pi],  \varphi \in [0, 2\pi]\right\} ~\land~ \hat{a} = 1 \Rightarrow \omega = 0.
\end{equation}

\noindent This provides an example of a family of bandlimited functions where the linear interpolation operation leads to a gain error. In addition, due to the Fourier representation theorem, it is always possible to find a base for $F$ in $\left\{sin(\omega x + \varphi) | \omega \in [0,\pi],  \varphi \in [0, 2\pi]\right\}$.

By virtue of proposition~\ref{prop:sumbiasgainformula}, it also mean that $\hat{a} \neq 1$ for almost all bandlimited functions, \ie given the functions $f_1$, $f_2$ and their approximation $\hat{f}_1$, $\hat{f}_2$ the measure of the space of approximation functions for which the optimal sum approximation will be of the form $\hat{f}_1 + \hat{f}_2 + cst$ has measure 0 (for the usual measure of $\Reals^2$, \ie the area, used on $L_{1px}$).

\end{proof}

\begin{proposition}
\label{prop:linearerrorbias}

Given an image signal similar to the one described in Proposition~\ref{prop:linearerrorcorrelation_base}, as well as its piecewise linear approximation, it is possible that $b \neq 0$.
\end{proposition}

\begin{proof}
If we take the solution provided in Equation~\eqref{eq:estsinsol} for $f(x) = sin(\pi x)$ we get:

\begin{equation}
    \hat{b} = \dfrac{2}{\pi} \neq 0.
\end{equation}

\noindent which demonstrates that linear interpolation can also add a bias when interpolating in a given pixel interval. Now, on average, by a simple symmetry argument (taking the negative of the signal gives the same bias with switched sign), we expect this bias to be $0$, but this would be only on average, as demonstrated here, for specific cases, the bias can be different than 0.

\end{proof}

Proposition~\ref{prop:linearerrorcorrelation_base} and Proposition~\ref{prop:linearerrorbias} show that a ZNCC-based closed form formula is expected to perform better than either of NCC, (Z)SSD, or (Z)SAD closed form formulae, even if the images used for patch-based matching have the same gain and bias (or at least the same bias). This is because linear interpolation itself can introduce a constant and/or a proportional error in the signal, and that ZNCC is not affected by either the constant or the proportional error caused by the linear interpolation procedure. This observation is supported by our experimental results, as ZNCC offered the best performance for image-based interpolation irrespective of the tested correlation window size.

\section{Proof of equivalence between the algebraic and geometric formula for (Z)NCC feature based refinement for 2 feature vectors}

Given two feature vectors of the target feature volume $f_{t,0}$ and $f_{t,1}$ (sitting on the same scanline, one pixel apart) and one feature vector from the source feature volume $f_{s}$, then the projected source feature vector $f_{s\perp}$ is given by (by expending equation 33):

\begin{equation}
   f_{s\perp} = \dfrac{ \dotprod{f_{s}}{f_{t,0}}\dotprod{f_{t,1}}{f_{t,1}} - \dotprod{f_{s}}{f_{t,1}}\dotprod{f_{t,1}}{f_{t,0}} }{ \dotprod{f_{t,0}}{f_{t,0}}\dotprod{f_{t,1}}{f_{t,1}}  - \dotprod{f_{t,0}}{f_{t,1}}\dotprod{f_{t,1}}{f_{t,0}} } f_{t,0} + 
   \dfrac{ \dotprod{f_{s}}{f_{t,1}}\dotprod{f_{t,0}}{f_{t,0}} - \dotprod{f_{s}}{f_{t,0}}\dotprod{f_{t,0}}{f_{t,1}} }{ \dotprod{f_{t,0}}{f_{t,0}}\dotprod{f_{t,1}}{f_{t,1}}  - \dotprod{f_{t,0}}{f_{t,1}}\dotprod{f_{t,1}}{f_{t,0}} } f_{t,1}
\end{equation}

Then, the vector $f_{t\perp}$ (Eq.34) is given by:

\begin{equation}
    f_{t\perp} = \dfrac{\dotprod{f_{t,0}-f_{t,1}}{-f_{t,1}}}{\dotprod{f_{t,0}-f_{t,1}}{f_{t,0}-f_{t,1}}} (f_{t,0}-f_{t,1}) + f_{t,1}
\end{equation}

Finally, by expending equations 35 and 36, expending each sum or difference withing the inner products and replacing each inner product $\dotprod{f_{t,0}}{f_{t,1}}$ by $\left\|f_{t,0}\right\|_2\left\|f_{t,1}\right\|_2 cos(\widehat{f_{t,0},0,f_{t,1}})$, where $\widehat{f_{t,0},0,f_{t,1}}$ is the unknown but constant angle between the target feature vectors, one obtain a rational function for $\hat{\alpha}$. From $\hat{\alpha}$, the expression for $\Delta \hat{d}$ is trivial to obtain from equation~32:

\begin{equation}
    \Delta \hat{d} = 1-\hat{\alpha}
\end{equation}

The geometric rational expression for $\hat{\alpha}$ can be compared to the one derived analytically (Eq.~21) which can also be expanded to use a scalar formulation:

\begin{equation}
\label{final_expr}
    \Delta \hat{d} = \dfrac{\dotprod{f_s}{f_{t,0}}\dotprod{f_{t,0}}{f_{t,1}} - \dotprod{f_s}{f_{t,1}}\dotprod{f_{t,0}}{f_{t,0}}}{
    \dotprod{f_s}{f_{t,0}}\dotprod{f_{t,0}}{f_{t,1}} - \dotprod{f_s}{f_{t,0}}\dotprod{f_{t,1}}{f_{t,1}}  - \dotprod{f_s}{f_{t,1}}\dotprod{f_{t,0}}{f_{t,0}} +
    \dotprod{f_s}{f_{t,1}}\dotprod{f_{t,0}}{f_{t,1}}
    }
\end{equation}

Expanding, simplifying and comparing those expressions is trivial but extremely tedious. So we did it using the computer algebra system maxima \cite{li2008maxima}, checking that both expressions are indeed equivalent.

\section{Additional results with the Middlebury mobile dataset and Active-Passive SimStereo.}

In addition to the main Middlebury dataset \cite{scharstein2014high}, which we used for our main experiment, we also evaluated the 1D refinement formulae on the more recent Middlebury mobile dataset \cite{middlebury2021}, as well as on the Active-Passive SimStereo Dataset \cite{NEURIPS2022_bc3a68a2}. It is important to underline that while the Middlebury mobile dataset is more recent, its focus is on mobile computation, and as such it used only a subset of the pipeline used to obtain the ground truth of the main dataset \cite{scharstein2014high}. As such, we can expect the measure to be slightly less precise. Active-Passive SimStereo on the other hand is a simulated dataset, meaning it contains basically no noise both in the images themselves and in the ground truth.

\begin{table*}[t]
\centering
\begin{adjustbox}{max width=\textwidth}
\begin{tabular}{ll|rrrrrr|rrrrr|}
\cline{3-13}
 &  & \multicolumn{6}{c|}{Mean MAE over Middlebury Mobile {[}px{]}} & \multicolumn{5}{c|}{Mean SNR over Middlebury Mobile {[}dB{]}} \\ \hline
\multicolumn{1}{|l|}{$C$} & $W$ & \multicolumn{1}{c}{Raw} & \multicolumn{1}{c}{Parabola} & \multicolumn{1}{c}{Equiangular} & \multicolumn{1}{c}{\cite{Shimizu2005}} & \multicolumn{1}{c}{Barycentric} & \multicolumn{1}{c|}{Predictive} & \multicolumn{1}{c}{Parabola} & \multicolumn{1}{c}{Equiangular} & \multicolumn{1}{c}{\cite{Shimizu2005}} & \multicolumn{1}{c}{Barycentric} & \multicolumn{1}{c|}{Predictive} \\ \hline

\multicolumn{1}{|l|}{NCC}           & 5x5                 & 0.28 & 0.19     & 0.20          & 0.19      & \textbf{0.17}                        & 0.18       & -15.95   & \textbf{-21.16} & -20.49          & -20.85          & -17.27      \\
\multicolumn{1}{|l|}{NCC}           & 7x7                 & 0.29 & 0.19     & 0.19          & 0.18      & \textbf{0.17}                        & 0.18       & -17.27   & -21.60          & -21.34          & \textbf{-21.99} & -17.81      \\
\multicolumn{1}{|l|}{NCC}           & 9x9                 & 0.30 & 0.19     & 0.19          & 0.19      & \textbf{0.18}                        & 0.19       & -18.46   & -21.29          & \textbf{-21.77} & -21.64          & -18.24      \\
\multicolumn{1}{|l|}{NCC}           & 11x11               & 0.30 & 0.19     & 0.19          & 0.19      & \textbf{0.19}                        & 0.19       & -19.23   & -20.45          & \textbf{-21.66} & -20.66          & -18.41      \\ \hline
\multicolumn{1}{|l|}{ZNCC}          & 5x5                 & 0.27 & 0.19     & 0.20          & 0.18      & {\color[HTML]{006600} \textbf{0.16}} & 0.19       & -15.35   & \textbf{-20.95} & -19.88          & -20.52          & -16.43      \\
\multicolumn{1}{|l|}{ZNCC}          & 7x7                 & 0.29 & 0.19     & 0.19          & 0.18      & \textbf{0.17}                        & 0.19       & -16.77   & \textbf{-22.36} & -21.42          & -22.30          & -17.04      \\
\multicolumn{1}{|l|}{ZNCC}          & 9x9                 & 0.30 & 0.19     & 0.19          & 0.18      & \textbf{0.17}                        & 0.19       & -18.23   & -21.73          & \textbf{-22.00} & -21.58          & -17.44      \\
\multicolumn{1}{|l|}{ZNCC}          & 11x11               & 0.30 & 0.19     & 0.19          & 0.19      & \textbf{0.18}                        & 0.19       & -18.97   & -20.58          & \textbf{-21.76} & -20.67          & -17.99      \\ \hline
\multicolumn{1}{|l|}{SSD}           & 5x5                 & 0.30 & 0.22     & 0.22          & 0.22      & \textbf{0.19}                        & 0.21       & -18.37   & -20.15          & -20.13          & \textbf{-20.34} & -16.03      \\
\multicolumn{1}{|l|}{SSD}           & 7x7                 & 0.31 & 0.21     & 0.22          & 0.21      & \textbf{0.19}                        & 0.21       & -18.85   & -19.78          & -20.13          & \textbf{-20.60} & -16.63      \\
\multicolumn{1}{|l|}{SSD}           & 9x9                 & 0.31 & 0.21     & 0.22          & 0.21      & \textbf{0.20}                        & 0.22       & -18.87   & -19.56          & -20.22          & \textbf{-20.49} & -16.89      \\
\multicolumn{1}{|l|}{SSD}           & 11x11               & 0.32 & 0.21     & 0.22          & 0.22      & \textbf{0.21}                        & 0.22       & -19.10   & -19.32          & -20.28          & \textbf{-20.62} & -17.01      \\ \hline
\multicolumn{1}{|l|}{ZSSD}          & 5x5                 & 0.28 & 0.19     & 0.19          & 0.20      & \textbf{0.17}                        & 0.18       & -15.91   & -21.19          & -19.27          & \textbf{-20.98} & -16.99      \\
\multicolumn{1}{|l|}{ZSSD}          & 7x7                 & 0.29 & 0.19     & 0.19          & 0.20      & \textbf{0.17}                        & 0.18       & -17.35   & -21.67          & -20.71          & \textbf{-22.46} & -17.61      \\
\multicolumn{1}{|l|}{ZSSD}          & 9x9                 & 0.30 & 0.19     & 0.19          & 0.20      & \textbf{0.18}                        & 0.19       & -18.49   & -21.00          & -21.05          & \textbf{-21.72} & -17.86      \\
\multicolumn{1}{|l|}{ZSSD}          & 11x11               & 0.30 & 0.19     & 0.19          & 0.21      & \textbf{0.19}                        & 0.19       & -19.32   & -20.23          & \textbf{-21.20} & -20.66          & -18.07      \\ \hline
\multicolumn{1}{|l|}{SAD}           & 5x5                 & 0.31 & 0.22     & 0.22          & 0.21      & \textbf{0.21}                        & 0.23       & -16.39   & -19.53          & -19.82          & \textbf{-20.37} & -17.17      \\
\multicolumn{1}{|l|}{SAD}           & 7x7                 & 0.31 & 0.22     & 0.21          & 0.21      & \textbf{0.21}                        & 0.22       & -17.18   & -20.23          & -20.08          & \textbf{-20.59} & -17.45      \\
\multicolumn{1}{|l|}{SAD}           & 9x9                 & 0.32 & 0.22     & 0.21          & 0.21      & \textbf{0.21}                        & 0.23       & -17.28   & -20.29          & -20.33          & \textbf{-20.47} & -18.09      \\
\multicolumn{1}{|l|}{SAD}           & 11x11               & 0.32 & 0.22     & \textbf{0.21} & 0.22      & 0.21                                 & 0.23       & -17.52   & -20.20          & -20.18          & \textbf{-20.46} & -18.09      \\ \hline
\multicolumn{1}{|l|}{ZSAD}          & 5x5                 & 0.28 & 0.19     & 0.19          & 0.20      & \textbf{0.18}                        & 0.20       & -12.99   & -17.92          & -18.07          & \textbf{-20.55} & -17.90      \\
\multicolumn{1}{|l|}{ZSAD}          & 7x7                 & 0.29 & 0.19     & 0.18          & 0.20      & \textbf{0.18}                        & 0.19       & -14.45   & -20.32          & -19.78          & \textbf{-22.26} & -18.58      \\
\multicolumn{1}{|l|}{ZSAD}          & 9x9                 & 0.30 & 0.19     & \textbf{0.18} & 0.20      & 0.18                                 & 0.19       & -15.60   & -21.29          & -20.28          & \textbf{-22.35} & -19.11      \\
\multicolumn{1}{|l|}{ZSAD}          & 11x11               & 0.30 & 0.19     & \textbf{0.19} & 0.20      & 0.19                                 & 0.19       & -16.33   & -21.18          & -20.27          & \textbf{-21.38} & -19.00      \\ \hline

\end{tabular}
\end{adjustbox}
\caption{Aggregate results of studied formulae for the Middlebury Mobile stereo dataset (1D patch-based matching) for different correlation windows and cost functions. The best performing formula per row for a given metric is bold and the best performing formula overall is in green.}
\label{table:results1dmobile}
\end{table*}

\begin{table*}[t]
\centering
\begin{adjustbox}{max width=\textwidth}

\begin{tabular}{ll|rrrrrr|rrrrr|}
\cline{3-13}
                                   &                     & \multicolumn{6}{l|}{Mean MAE over Active-Passive SimStereo {[}px{]}}                                                                                                                               & \multicolumn{5}{l|}{Mean SNR over Active-Passive SimStereo {[}dB{]}}                                                                                         \\ \hline
\multicolumn{1}{|l}{$C$} & $W$ & \multicolumn{1}{l}{Raw} & \multicolumn{1}{l}{Parabola} & \multicolumn{1}{l}{Equiangular} & \multicolumn{1}{l}{\cite{Shimizu2005}} & \multicolumn{1}{l}{Barycentric}             & \multicolumn{1}{l|}{Predictive} & \multicolumn{1}{l}{Parabola} & \multicolumn{1}{l}{Equiangular} & \multicolumn{1}{l}{\cite{Shimizu2005}} & \multicolumn{1}{l}{Barycentric} & \multicolumn{1}{l|}{Predictive} \\ \hline
\multicolumn{1}{|l}{NCC} & 5x5 & 0.26  & 0.14 & 0.15  & 0.14  & \textbf{0.13} & 0.13  & -15.93  & -20.60 & \textbf{-20.84} & -20.19 & -13.81 \\
\multicolumn{1}{|l}{NCC}  & 7x7  & 0.28  & 0.14  & 0.14  & 0.13  & \textbf{0.13}  & 0.14  & -17.33  & -19.01  & \textbf{-21.43}  & -18.33  & -14.69 \\
\multicolumn{1}{|l}{NCC} & 9x9  & 0.29  & 0.14  & 0.14 & \textbf{0.13} & 0.14 & 0.14 & -17.99 & -17.94 & \textbf{-21.45} & -16.96 & -15.72 \\
\multicolumn{1}{|l}{NCC} & 11x11 & 0.29 & 0.15 & 0.15 & \textbf{0.14} & 0.15  & 0.15 & -18.48  & -17.34 & \textbf{-21.43} & -16.27 & -16.62 \\ \hline
\multicolumn{1}{|l}{ZNCC} & 5x5  & 0.26 & 0.15 & 0.15  & 0.13 & {\color[HTML]{006600} \textbf{0.12}} & 0.14 & -15.41 & -20.90 & \textbf{-20.43} & -19.80 & -13.22 \\
\multicolumn{1}{|l}{ZNCC} & 7x7 & 0.28 & 0.14  & 0.14  & \textbf{0.13} & \textbf{0.13} & 0.14 & -16.84 & -19.47 & \textbf{-21.20}  & -18.24 & -13.92 \\
\multicolumn{1}{|l}{ZNCC}  & 9x9  & 0.28 & 0.14 & 0.14  & \textbf{0.13} & 0.13 & 0.14  & -17.59 & -18.39 & \textbf{-21.37}  & -16.85 & -14.85 \\
\multicolumn{1}{|l}{ZNCC} & 11x11 & 0.29 & 0.15 & 0.15 & \textbf{0.13} & 0.14 & 0.14 & -18.17 & -17.71 & \textbf{-21.42} & -16.20 & -15.71 \\ \hline
\multicolumn{1}{|l}{SSD} & 5x5   & 0.28 & 0.15 & 0.16 & 0.15 & \textbf{0.14}  & 0.15 & -18.53 & -18.55  & \textbf{-20.71} & -18.46 & -14.84 \\
\multicolumn{1}{|l}{SSD} & 7x7 & 0.29  & 0.15 & 0.15 & \textbf{0.14}  & 0.14 & 0.15  & -18.65 & -17.26 & \textbf{-20.56} & -16.83 & -15.33 \\
\multicolumn{1}{|l}{SSD} & 9x9 & 0.29 & 0.15 & 0.15 & \textbf{0.14}  & 0.15 & 0.15 & -18.73 & -16.64 & \textbf{-20.41} & -16.08 & -16.04 \\
\multicolumn{1}{|l}{SSD}  & 11x11 & 0.30 & 0.16 & 0.16 & \textbf{0.15} & 0.16 & 0.16 & -18.90 & -16.26 & \textbf{-20.40} & -15.57 & -16.68 \\ \hline
\multicolumn{1}{|l}{ZSSD} & 5x5 & 0.26 & 0.15 & 0.15 & 0.13 & \textbf{0.13} & 0.14 & -15.87  & \textbf{-20.58} & -20.52  & -20.23  & -13.93 \\
\multicolumn{1}{|l}{ZSSD} & 7x7 & 0.28 & 0.14 & 0.14 & \textbf{0.13} & 0.13 & 0.14 & -17.25 & -19.16 & \textbf{-21.24} & -18.46  & -14.79 \\
\multicolumn{1}{|l}{ZSSD}  & 9x9  & 0.29 & 0.14  & 0.14  & \textbf{0.13}                & 0.14 & 0.14  & -18.00 & -18.14 & \textbf{-21.46} & -17.16 & -15.81 \\
\multicolumn{1}{|l}{ZSSD} & 11x11 & 0.29 & 0.15 & 0.15 & \textbf{0.14} & 0.15 & 0.15 & -18.58 & -17.48 & \textbf{-21.55} & -16.41 & -16.66 \\ \hline
\multicolumn{1}{|l}{SAD} & 5x5 & 0.28 & 0.16 & 0.15 & \textbf{0.14} & 0.15 & 0.16 & -14.75 & -19.73 & -19.32 & \textbf{-19.95} & -16.16 \\
\multicolumn{1}{|l}{SAD} & 7x7 & 0.29 & 0.16 & 0.14 & \textbf{0.14} & 0.15 & 0.16 & -15.18 & \textbf{-19.47} & -19.26 & -18.54 & -16.41 \\
\multicolumn{1}{|l}{SAD} & 9x9& 0.29& 0.16 & 0.14  & \textbf{0.14} & 0.16 & 0.16 & -15.60 & -19.15 & \textbf{-19.28} & -17.56 & -17.10 \\
\multicolumn{1}{|l}{SAD} & 11x11 & 0.30 & 0.16 & 0.15 & \textbf{0.14} & 0.16 & 0.16 & -16.11 & -19.01 & \textbf{-19.40} & -16.98 & -17.70 \\ \hline
\multicolumn{1}{|l}{ZSAD} & 5x5 & 0.26 & 0.15 & 0.14 & \textbf{0.13} & 0.14 & 0.15 & -11.57 & -18.36 & -19.02 & \textbf{-20.95} & -15.34 \\
\multicolumn{1}{|l}{ZSAD} & 7x7 & 0.28 & 0.15 & 0.14 & \textbf{0.13} & 0.14 & 0.15 & -12.89 & \textbf{-19.74} & -19.67 & -19.73 & -16.04 \\
\multicolumn{1}{|l}{ZSAD} & 9x9  & 0.28 & 0.15 & 0.14 & \textbf{0.13} & 0.14 & 0.15 & -13.78 & \textbf{-19.96} & -19.88 & -18.56 & -17.03 \\
\multicolumn{1}{|l}{ZSAD} & 11x11 & 0.29 & 0.15 & 0.14 & \textbf{0.13} & 0.15 & 0.15 & -14.69 & \textbf{-20.16} & -20.13 & -17.78 & -17.90 \\ \hline
\end{tabular}
\end{adjustbox}
\caption{Aggregate results of studied formulae for the Active-Passive SimStereo (1D patch-based matching) for different correlation windows and cost functions. The best performing formula per row for a given metric is bold and the best performing formula overall is in green.}
\label{table:resultsactivepassivesimstereo}
\end{table*}

The results for the Middlebury mobile dataset, reported in Table~\ref{table:results1dmobile}, are overall coherent with the results we obtained on the main Middlebury dataset. Especially, the Barycentric feature interpolation method with the ZNCC cost function and a $5 \times 5$ windows still yields the best accuracy among all considered methods. The accuracy is lower, but this can be attributed to the fact that the ground truth for the Middlebury mobile dataset is known to be less accurate than in the Middlebury 2014 dataset.

The same goes for the Active-Passive SimStereo results, reported in Table~\ref{table:resultsactivepassivesimstereo}, which are globally coherent with the results obtained with the other dataset, with the key difference that the method of Shimizu and Okutomi~\cite{2DSimultaneousSupixel} performs better on this dataset. Nonetheless, the top performing formula is still ours, based on image interpolation, with a small $5 \times 5$ windows and the ZNCC cost function. The overall gain in accuracy for all methods can probably be attributed to the lack of noise both in the images and the ground truth disparity.

The fact that the proposed formula yield the highest accuracy for three different datasets with different level of noise in their ground truth disparities provide strong confidence in the significance of the observed results.

\section{Comparison between Linear and spline interpolation}

\begin{table*}[bt]
\centering
\begin{adjustbox}{max width=\textwidth}
\begin{tabular}{|ll|rrrrrr|}
\hline
\textbf{Interpolation method} & \textbf{Window size} & \multicolumn{6}{l|}{\textbf{MAE per Cost functions}}      \\ \hline
                              &                      & NCC    & ZNCC  & SSD    & ZSSD  & SAD    & ZSAD   \\ \hline
\textbf{Middlebury 2014}     & \textbf{5x5}         &        &       &        &       &        &        \\ \hline
Linear                        &                      & 0.134  & 0.124 & 0.164  & 0.133 & 0.184  & 0.142  \\
Spline                        &                      & 0.133  & 0.124 & 0.161  & 0.133 & 0.181  & 0.140  \\
Delta                         &                      & -0.001 & 0.000 & -0.003 & 0.000 & -0.003 & -0.002 \\ \hline
\textbf{}                     & \textbf{7x7}         &        &       &        &       &        &        \\ \hline
Linear                        &                      & 0.141  & 0.131 & 0.166  & 0.140 & 0.188  & 0.146  \\
Spline                        &                      & 0.141  & 0.132 & 0.164  & 0.141 & 0.185  & 0.145  \\
Delta                         &                      & 0.000  & 0.001 & -0.002 & 0.001 & -0.003 & -0.001 \\ \hline
\textbf{}                     & \textbf{9x9}         &        &       &        &       &        &        \\ \hline
Linear                        &                      & 0.151  & 0.141 & 0.172  & 0.151 & 0.192  & 0.154  \\
Spline                        &                      & 0.152  & 0.143 & 0.171  & 0.152 & 0.189  & 0.153  \\
Delta                         &                      & 0.001  & 0.002 & -0.001 & 0.001 & -0.003 & -0.001 \\ \hline
\textbf{}                     & \textbf{11x11}       &        &       &        &       &        &        \\ \hline
Linear                        &                      & 0.163  & 0.154 & 0.181  & 0.164 & 0.197  & 0.163  \\
Spline                        &                      & 0.165  & 0.156 & 0.180  & 0.166 & 0.196  & 0.164  \\
Delta                         &                      & 0.002  & 0.002 & -0.001 & 0.002 & -0.001 & -0.001 \\ \hline
\end{tabular}
\end{adjustbox}
\caption{Comparisons of the accuracy of linear and spline based features interpolation for disparity refinement in the Middlebury 2014 dataset.}
\label{table:resultssplines}
\end{table*}

The closed form formulae that were presented in this work assume that linear interpolation is used to interpolate the features. An interesting question one might ask, is how much accuracy is lost by using linear interpolation instead of, for example, spline interpolation. Keys has shown that a third-order convergence interpolation spline approximation of a signal can be obtained via convolution with a piecewise polynomial kernel \cite{1163711}, yielding the standard cubic-convolution algorithm. In this section, we use bi-cubic interpolation to reconstruct the feature volume at a fine resolution (1/10th). We then obtain an estimate of the disparity using the linear-interpolation based method. While time consuming, this approach avoid the caveats that gradient descent methods can have, especially convergence to local minima or failure to converge, so it is better for evaluation purposes.

The results, outlined in Table~\ref{table:resultssplines}, show that there is really very little difference between using linear interpolation and spline interpolation. While this might seem surprising at first, as splines offer a much smoother interpolation, one has to remember that the goal here is not to interpolate the features as closely as possible, but to locate the subpixel minimal cost (or maximal score). In fact, as the performances of the parabola and equiangular interpolation kernel were somehow similar for cost interpolation, it is not so surprising to see that splines and linear interpolation yield similar performance for image-based subpixel disparity interpolation.

\section{Analysis per images}

While results averaged out over a whole dataset offer a good overview of the performance of each formula, it is important to check if the results are significant or not, \ie if the best performing formula yield better results for less than 50\% of images in the dataset and is the best only because it performed really well on a single image, then it is not really advantageous. To check that we have measured, for both the Middlebury 2014 \cite{scharstein2014high} and Active-Passive SimStereo \cite{NEURIPS2022_bc3a68a2} datasets, the proportion of images for which each method yielded the best results, compared to other methods on the same images.

\begin{table*}[t]
\centering
\begin{adjustbox}{max width=\textwidth}
\begin{tabular}{ll|rrrrrrrrrr|}
\cline{3-12}
                                   &              & \multicolumn{10}{l|}{Percent of images the method yield the best results compared to other method}                                                                                                                                                                                                                                      \\ \cline{3-12} 
                                   &              & \multicolumn{5}{l|}{Middlebury 2014}                                                                                                                              & \multicolumn{5}{l|}{Active-Passive SimStereo dataset}                                                                                                              \\ \hline
\multicolumn{1}{|l}{$C$} & $W$ & \multicolumn{1}{l}{Parabola} & \multicolumn{1}{l}{Equiangular} & \multicolumn{1}{l}{\cite{Shimizu2005}} & \multicolumn{1}{l}{Barycentric} & \multicolumn{1}{l|}{Predictive} & \multicolumn{1}{l}{Parabola} & \multicolumn{1}{l}{Equiangular} & \multicolumn{1}{l}{\cite{Shimizu2005}} & \multicolumn{1}{l}{Barycentric} & \multicolumn{1}{l|}{Predictive} \\ \hline
\multicolumn{1}{|l}{NCC} & 5x5 & 0\% & 0\% & 0\% & 100\% & \multicolumn{1}{r|}{0\%} & 1\% & 0\% & 8\% & 83\% & 8\%\\
\multicolumn{1}{|l}{NCC} & 7x7 & 0\% & 10\% & 0\% & 90\% & \multicolumn{1}{r|}{0\%} & 0\% & 0\% & 30\% & 59\% & 11\%\\
\multicolumn{1}{|l}{NCC} & 9x9 & 0\% & 10\% & 30\% & 60\% & \multicolumn{1}{r|}{10\%} & 0\% & 1\% & 61\% & 32\% & 6\%\\
\multicolumn{1}{|l}{NCC} & 11x11 & 0\% & 10\% & 50\% & 40\% & \multicolumn{1}{r|}{10\%} & 1\% & 1\% & 79\% & 19\% & 0\%\\\hline
\multicolumn{1}{|l}{ZNCC} & 5x5 & 0\% & 0\% & 0\% & 100\% & \multicolumn{1}{r|}{0\%} & 1\% & 0\% & 17\% & 82\% & 0\%\\
\multicolumn{1}{|l}{ZNCC} & 7x7 & 0\% & 0\% & 0\% & 100\% & \multicolumn{1}{r|}{0\%} & 0\% & 0\% & 45\% & 53\% & 2\%\\
\multicolumn{1}{|l}{ZNCC} & 9x9 & 0\% & 0\% & 20\% & 70\% & \multicolumn{1}{r|}{10\%} & 0\% & 0\% & 69\% & 30\% & 1\%\\
\multicolumn{1}{|l}{ZNCC} & 11x11 & 0\% & 0\% & 40\% & 50\% & \multicolumn{1}{r|}{10\%} & 0\% & 0\% & 77\% & 22\% & 1\%\\\hline
\multicolumn{1}{|l}{SSD} & 5x5 & 0\% & 0\% & 0\% & 100\% & \multicolumn{1}{r|}{0\%} & 0\% & 0\% & 16\% & 78\% & 7\%\\
\multicolumn{1}{|l}{SSD} & 7x7 & 0\% & 0\% & 10\% & 90\% & \multicolumn{1}{r|}{0\%} & 0\% & 0\% & 44\% & 50\% & 6\%\\
\multicolumn{1}{|l}{SSD} & 9x9 & 10\% & 10\% & 0\% & 80\% & \multicolumn{1}{r|}{0\%} & 1\% & 0\% & 71\% & 27\% & 1\%\\
\multicolumn{1}{|l}{SSD} & 11x11 & 20\% & 20\% & 10\% & 50\% & \multicolumn{1}{r|}{0\%} & 1\% & 0\% & 85\% & 13\% & 1\%\\\hline
\multicolumn{1}{|l}{ZSSD} & 5x5 & 0\% & 0\% & 0\% & 100\% & \multicolumn{1}{r|}{0\%} & 0\% & 0\% & 33\% & 67\% & 0\%\\
\multicolumn{1}{|l}{ZSSD} & 7x7 & 0\% & 10\% & 0\% & 90\% & \multicolumn{1}{r|}{0\%} & 0\% & 0\% & 55\% & 44\% & 1\%\\
\multicolumn{1}{|l}{ZSSD} & 9x9 & 10\% & 20\% & 0\% & 70\% & \multicolumn{1}{r|}{0\%} & 0\% & 1\% & 73\% & 24\% & 2\%\\
\multicolumn{1}{|l}{ZSSD} & 11x11 & 10\% & 40\% & 0\% & 50\% & \multicolumn{1}{r|}{0\%} & 0\% & 2\% & 85\% & 11\% & 2\%\\\hline
\multicolumn{1}{|l}{SAD} & 5x5 & 0\% & 60\% & 20\% & 20\% & \multicolumn{1}{r|}{0\%} & 0\% & 0\% & 72\% & 28\% & 0\%\\
\multicolumn{1}{|l}{SAD} & 7x7 & 0\% & 100\% & 0\% & 0\% & \multicolumn{1}{r|}{0\%} & 0\% & 1\% & 94\% & 5\% & 0\%\\
\multicolumn{1}{|l}{SAD} & 9x9 & 0\% & 100\% & 0\% & 0\% & \multicolumn{1}{r|}{0\%} & 0\% & 5\% & 92\% & 3\% & 0\%\\
\multicolumn{1}{|l}{SAD} & 11x11 & 10\% & 90\% & 0\% & 0\% & \multicolumn{1}{r|}{0\%} & 0\% & 8\% & 91\% & 1\% & 0\%\\\hline
\multicolumn{1}{|l}{ZSAD} & 5x5 & 0\% & 30\% & 0\% & 70\% & \multicolumn{1}{r|}{0\%} & 0\% & 1\% & 69\% & 30\% & 0\%\\
\multicolumn{1}{|l}{ZSAD} & 7x7 & 0\% & 50\% & 0\% & 50\% & \multicolumn{1}{r|}{0\%} & 0\% & 3\% & 85\% & 12\% & 0\%\\
\multicolumn{1}{|l}{ZSAD} & 9x9 & 0\% & 90\% & 0\% & 10\% & \multicolumn{1}{r|}{0\%} & 0\% & 6\% & 89\% & 5\% & 0\%\\
\multicolumn{1}{|l}{ZSAD} & 11x11 & 0\% & 90\% & 0\% & 10\% & \multicolumn{1}{r|}{0\%} & 0\% & 6\% & 92\% & 1\% & 1\%\\\hline
\end{tabular}
\end{adjustbox}
\caption{Proportion of images in different datasets where each formula yielded the best estimate of the refined disparity.}
\label{table:results_significance}
\end{table*}

The results, reported in Table~\ref{table:results_significance} show that when the reported overall accuracy clearly shows that a formula is better (\ie more than 0.005px improvement), then the results are significants. Our proposed formula, for the best configuration (a $5 \times 5$ windows and the ZNCC cost function), yield the best estimate for 100\% of the Middlebury images and 82\% for the Active-Passive SimStereo images. This provide some insight that, while there is a bit of variability, the proposed formula systematically outperform the other methods in the optimal configuration.

\bibliographystyle{ieee_fullname}
\bibliography{bibliography}